\documentclass{article}
\usepackage{iclr2026_conference,times}

\usepackage{hyperref}       % hyperlinks
\usepackage{url}            % simple URL typesetting
\usepackage{booktabs}       % professional-quality tables
\usepackage{amsfonts}       % blackboard math symbols
\usepackage{nicefrac}       % compact symbols for 1/2, etc.
\usepackage{microtype}      % microtypography
\usepackage{xcolor}    
\usepackage{graphicx}% colors
\usepackage{subcaption}
\usepackage{algorithm}
\usepackage{algpseudocode}
\usepackage{graphicx}
\usepackage{subcaption}
\usepackage{amsmath}
\definecolor{lightgray}{gray}{0.95}
\title{OrtSAE: Orthogonal Sparse Autoencoders \\ Uncover Atomic Features}

\author{%
\hspace{.06em} Anton Korznikov\thanks{Correspondence to korznikovantona@gmail.com}, \hspace{.3em}\
Andrey Galichin, \hspace{.3em}\
Alexey Dontsov, \hspace{.3em}\\
\textbf{
Oleg Y. Rogov, \hspace{.3em}\
Elena Tutubalina\thanks{These authors contributed equally to this work.} \hspace{.3em} \& \
Ivan Oseledets\footnotemark[2]  \hspace{.3em}\
}\\
[1ex]
}

\iclrfinalcopy
\begin{document}

\maketitle

% These phenomena prevents the extraction of irreducible, atomic features that are critical for clear mechanistic insights.

\begin{abstract}
Sparse autoencoders (SAEs) are a technique for sparse decomposition of neural network activations into human-interpretable features. However, current SAEs suffer from feature absorption, where specialized features capture instances of general features creating representation holes, and feature composition, where independent features merge into composite representations. In this work, we introduce Orthogonal SAE (OrtSAE), a novel approach aimed to mitigate these issues by enforcing orthogonality between the learned features. By implementing a new training procedure that penalizes high pairwise cosine similarity between SAE features, OrtSAE promotes the development of disentangled features while scaling linearly with the SAE size, avoiding significant computational overhead. We train OrtSAE across different models and layers and compare it with other methods. We find that OrtSAE discovers $9\%$ more distinct features, reduces feature absorption (by $65\%$) and composition (by $15\%$), improves performance on spurious correlation removal ($+6\%$), and achieves on-par performance for other downstream tasks compared to traditional SAEs.
\end{abstract}

% We evaluate OrtSAE against other methods across a range of models and layers.

\section{Introduction}

% 1. LLMs -> Mech. Interp.
% 2. (Optional) Polysemanticity + Superposition
% 3. SAEs + problem?
% 4. Feature absorption & composition; make some hint to similarity? + Tell why it is actually bad (hurts interpretability, worse at downstream tasks)
% 5. We notice that ... ; so we propose ...

Large Language Models (LLMs) have achieved remarkable performance in natural language processing, but their internal mechanisms remain poorly understood. Mechanistic interpretability aims to understand how neural networks function by reverse-engineering their computational processes \citep{olah2020zoom}. Central to this field is understanding $\textit{features}$, the human-interpretable concepts represented as directions in a model's internal representation \citep{elhage2022toy, park2023linear}.

Early interpretability methods focused on analyzing individual neurons \citep{olah2020zoom, bills2023language}, but a key challenge has been that neurons are often \textit{polysemantic}, responding to multiple unrelated concepts rather than encoding single interpretable features \citep{olah2020zoom}. One theory of why polysemanticity occurs is \textit{superposition}, which posits that neural networks represent more features than they have dimensions \citep{elhage2022toy}. Although this enables efficient use of model capacity, it significantly complicates interpretability research.

% Sparse Autoencoders (SAEs) have emerged as a powerful approach to disentangling superposition \citep{bricken2023towards, cunningham2023sparse}. By adding a sparsity penalty to the reconstruction loss, SAEs learn to decompose activations into a sparse latent space where each dimension aims to capture a distinct, interpretable feature \citep{gao2024scaling, marks2024sparse}. Traditional SAE variants \citep{bricken2023towards, gao2024scaling, bussmann2024batchtopk, rajamanoharan2024jumping} focused on improving reconstruction quality while maintaining sparsity. In contrast, \citet{bussmann2025learning} introduced a new hierarchical approach to organizing features at multiple levels of abstraction. However, the standard objective can lead to two failure modes. As the number of SAE latents grows, \textit{feature absorption} can occur (Fig. \ref{fig:absorption}), where a broad feature representation absorbs into more specific, token-aligned latents (e.g. a latent ``starts with E'' will activate on all tokens starting with ``E'', except for the token ``elephant'') \citep{chanin2024absorption}. Another issue is \textit{feature composition} (Fig. \ref{fig:composition}), in which independent features (e.g. representing ``red'' and ``square'') are merged into a single composite feature (``red square'') \citep{leask2025sparse}. Both problems undermine the interpretability of SAE latents and the applicability of SAE representations for downstream tasks \citep{karvonen2025saebench}.

\begin{figure}[ht!]
    \centering
    \begin{subfigure}[b]{0.45\textwidth}
        \centering
        \includegraphics[width=\textwidth]{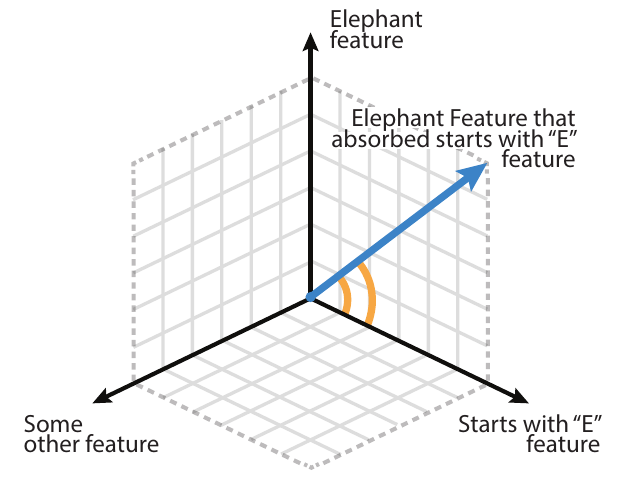}
        \caption{Example of feature absorption}
        \label{fig:absorption}
    \end{subfigure}
    \hfill
    \begin{subfigure}[b]{0.45\textwidth}
        \centering
        \includegraphics[width=\textwidth]{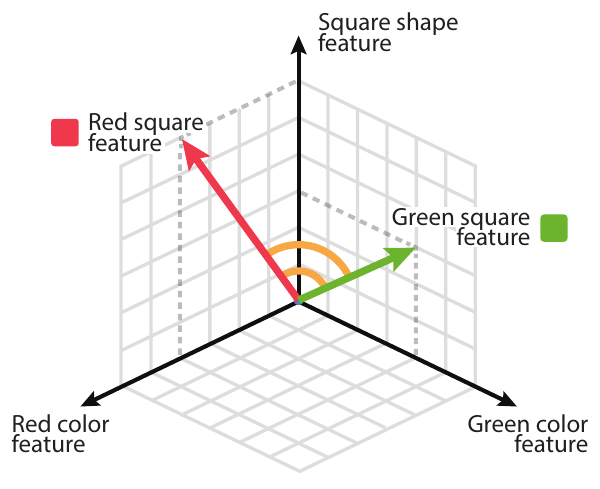}
        \caption{Example of feature composition}
        \label{fig:composition}
    \end{subfigure}
    \caption{\textbf{Illustration of feature absorption and feature composition}: (a) In feature absorption, specific features like ``elephant'' absorb broader features like ``starts with E''.  (b) In feature composition, independent concepts like ``red color'' and ``square form'' are merged into composite features.}
    \label{fig:composition_absorbtion}
\end{figure}

Sparse Autoencoders (SAEs) have emerged as a powerful approach to disentangling superposition \citep{bricken2023towards, cunningham2023sparse}. By adding a sparsity penalty to the reconstruction loss, SAEs learn to decompose activations into a sparse latent space where each dimension aims to capture a distinct, interpretable feature \citep{gao2024scaling, marks2024sparse}. Traditional SAE variants \citep{bricken2023towards, gao2024scaling, bussmann2024batchtopk, rajamanoharan2024jumping} focused on improving reconstruction quality while maintaining sparsity. However, the standard objective can lead to two failure modes. As the number of SAE latents grows, \textit{feature absorption} can occur (Fig. \ref{fig:absorption}), where a broad feature representation absorbs into more specific, token-aligned latents (e.g., a latent ``starts with E'' will activate on all tokens starting with ``E'', except for the token ``elephant'') \citep{chanin2024absorption}. Another issue is \textit{feature composition} (Fig. \ref{fig:composition}), in which independent features (e.g. representing ``red'' and ``square'') are merged into a single composite feature (``red square'') \citep{leask2025sparse}. Both problems undermine the interpretability of SAE latents and the applicability of SAE representations for downstream tasks \citep{karvonen2025saebench}. To address these issues, \citet{bussmann2025learning} introduced Matryoshka SAE, a hierarchical approach to organizing features at multiple levels of abstraction. However, this method introduces additional computational overhead and suffers from feature hedging \citep{chanin2025feature}, a problem where correlated features merge at higher levels, reducing interpretability. This highlights the need for alternative approaches.

\begin{figure}[t]
    \centering
    \begin{subfigure}[t]{0.71\textwidth}
        \centering
        \includegraphics[width=\textwidth]{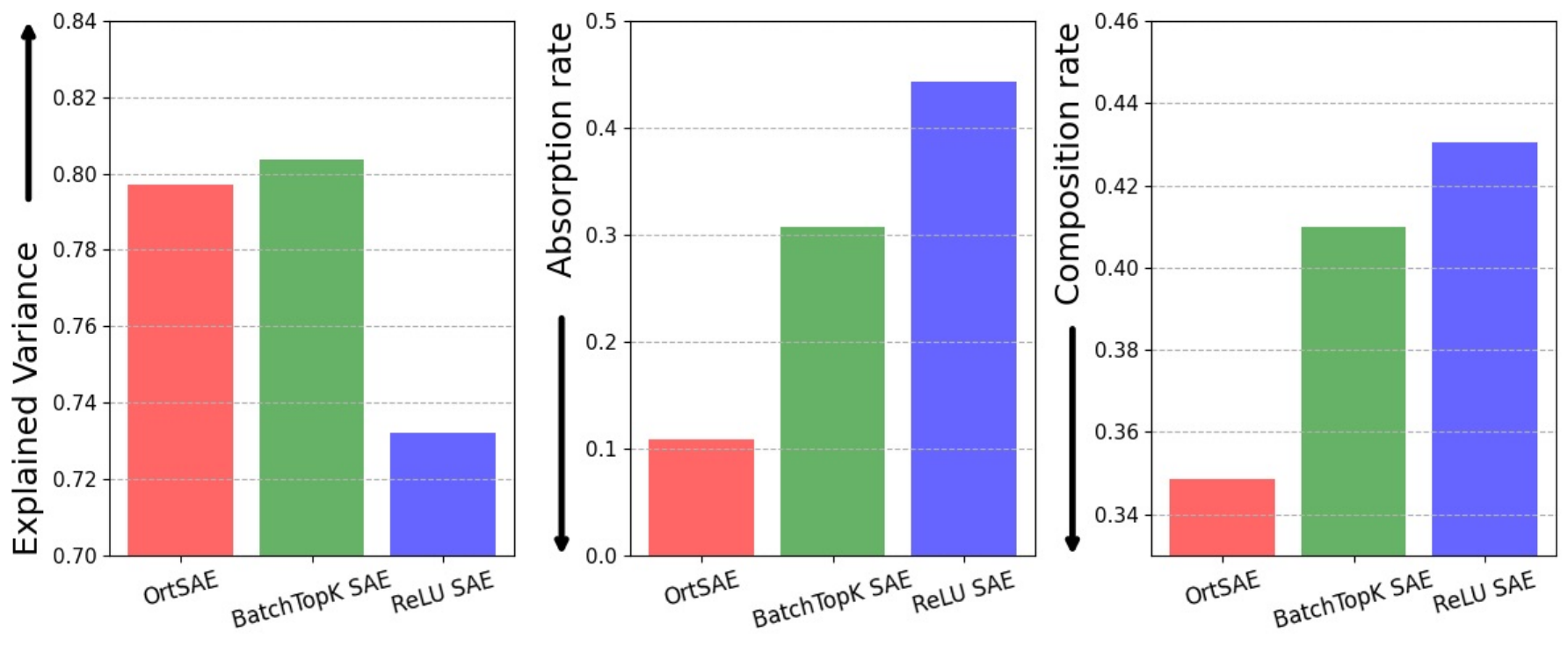}
        \label{fig:explained_variance}
    \end{subfigure}
        \begin{subfigure}[t]{0.284\textwidth}
        \centering
        \includegraphics[width=\textwidth]{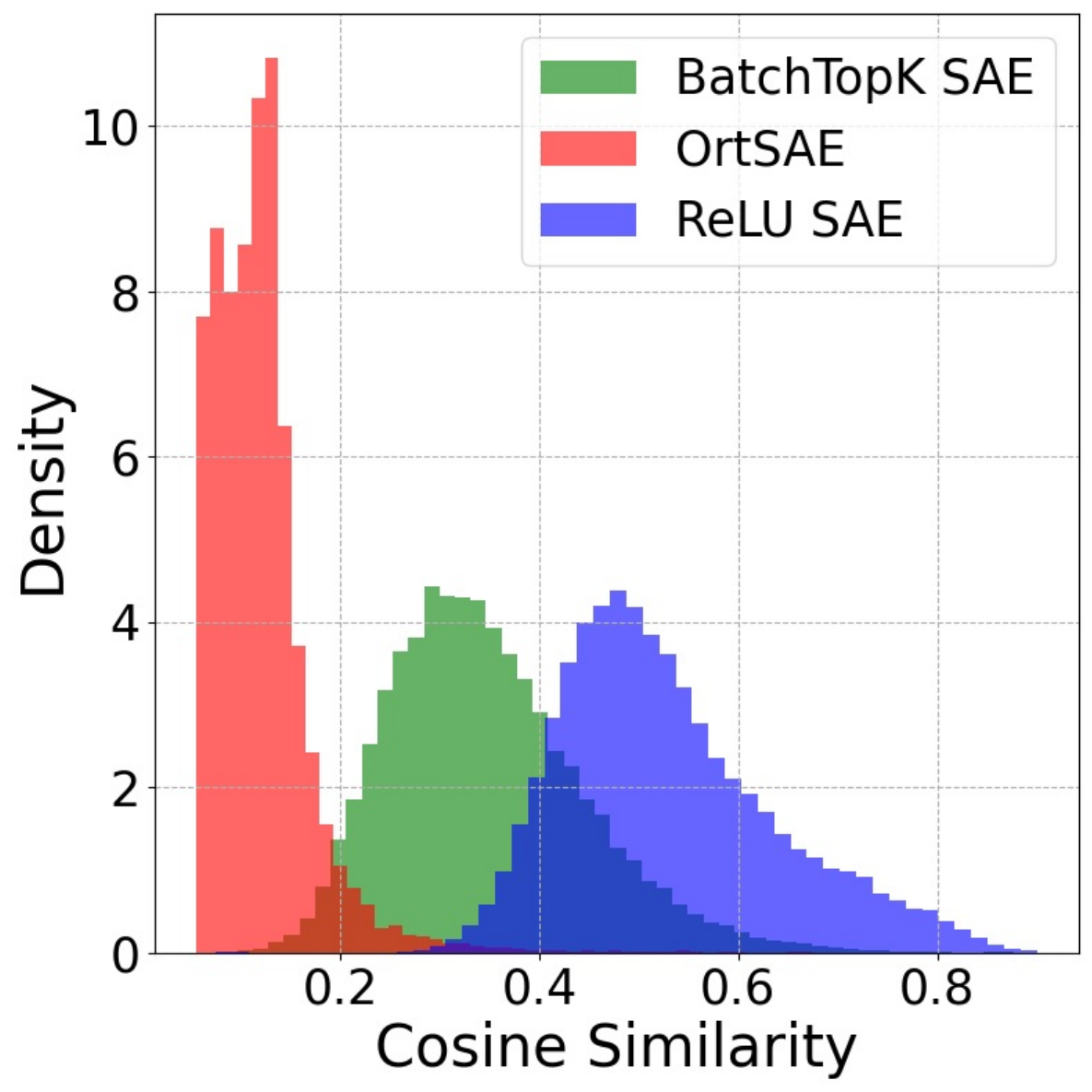}
        \label{fig:orthogonality}
    \end{subfigure}
    \caption{
  \textbf{Performance of OrtSAEs vs. traditional SAEs.} Bar plots display explained variance, absorption, and composition rates for three SAE variants at $\text{L}0$=70 sparsity. OrtSAEs show a marginally lower explained variance than BatchTopK SAEs but decreased absorption and composition, indicating better feature specificity. The density plot illustrates the distribution of pairwise cosine similarity values, computed as the maximum similarity between each decoder feature and its closest counterpart in the model, across all features at $\text{L}0$=70. OrtSAEs demonstrate lower pairwise cosine similarity, confirming greater decoder feature orthogonality compared to BatchTopK and ReLU SAEs.
  }
    \label{fig:main_metrics}
\end{figure}

Feature absorption and composition lead to redundant representations where multiple latents capture overlapping concepts, which results in high cosine similarities between them. This suggests that enforcing orthogonality between SAE latents could provide a principled approach to mitigate these issues.  Therefore, we propose \textbf{OrtSAE}, a novel approach to SAE training that promotes the emergence of more atomic features (Sec.~\ref{subsec:ortsae_training}). At each training step, we penalize high cosine similarities between SAE latents by introducing an additional orthogonality penalty. To optimize computation, we implement a chunk-wise strategy that divides SAE latents into smaller blocks, computes the penalty separately, and aggregates the results. This reduces the complexity from quadratic to linear with respect to the number of latents and introduces a negligible computational overhead. Importantly, this penalty scales efficiently without altering the core SAE architecture.

We train OrtSAE on the Gemma-2-2B \citep{team2024gemma} and Llama-3-8B \citep{dubey2024llama} and compare it against traditional SAEs and Matryoshka SAE \citep{bussmann2025learning}. Experimental results demonstrate that our objective reduces feature absorption and composition across a wide range of sparsity levels (Sec.~\ref{subsec:Atomicity investigation}). For example, at a $\text{L}0$ of $70$ (Fig.~\ref{fig:main_metrics}), OrtSAE discovers $9\%$ more distinct features, reduces feature absorption by $65\%$, and feature composition by $15\%$ compared to traditional SAEs. On SAEBench \citep{karvonen2025saebench}, our method improves performance on spurious correlation removal by $6\%$ while maintaining on-par performance for other downstream tasks (Sec.~\ref{subsec:Downstream benchmarks}). Through qualitative experiments, we show that OrtSAE features efficiently decompose composite features learned by other SAEs into more atomic components (Sec.~\ref{subsec:Atomicity investigation}). 

Our paper makes the following contributions:
\begin{itemize}
    \item We propose OrtSAE, a novel approach to SAE training that directly addresses the issues of feature absorption and composition, without requiring complex architectural changes or significant computational overhead (Sec.~\ref{subsec:ortsae_training}).
    \item Comparison of OrtSAE with traditional SAEs shows that our method produces more distinct features, reduces absorption and composition rates (Sec.~\ref{subsec:Atomicity investigation}).
    \item Experimental results on SAEBench demonstrate that our method performs on-par with other SAE architectures, and outperforms them on spurious correlation removal (Sec.~\ref{subsec:Downstream benchmarks}).
\end{itemize}

% \textbf{Organization of the paper.} In Section \ref{sec:related_work}, we discuss related work. In Section \ref{sec:Orthogonal SAE}, we provide details of the proposed method. The experimental settings and the results of our approach are discussed in Section \ref{sec:Experiments}. In Section \ref{sec:limitations}, we address the limitations of OrtSAE. We conclude the paper in Section \ref{sec:Discussion and Conclusion}.

\section{Related work}
\label{sec:related_work}

Sparse Autoencoders (SAEs) have gained significant traction for interpreting LLMs, addressing the challenge that LLMs often function as ``black boxes''. A key issue is polysemanticity, where individual neurons respond to multiple unrelated concepts \citep{bricken2023towards}. SAEs aim to resolve this by decomposing dense LLM activation vectors into a sparse set of monosemantic features, each representing a single concept \citep{cunningham2023sparse, huben2025sparse}. Pioneering work \citet{bricken2023towards} and \citet{cunningham2023sparse} demonstrated the effectiveness of this approach on small transformers, finding interpretable features such as DNA sequences or legal text. 
Subsequent efforts scaled SAEs to larger models like Claude 3 Sonnet \citep{templeton2024scaling} and GPT-4 \citep{gao2024scaling}, as well as open-source models \citep{lieberum-etal-2024-gemma}. 

% These features can be used to interpret attention mechanism \citep{kissane2024interpreting} and for circuits analysis \citep{marks2024sparse}. Beyond feature identification, SAEs are explored for analyzing model behaviors like hallucination or bias \citep{li2025unlearning} and steering model outputs by manipulating feature activations \citep{huben2025sparse}. 

Limitations in basic SAEs, typically using ReLU activation with an $\text{L}1$ penalty \citep{bricken2023towards}, such as $\text{L}1$-induced shrinkage (underestimation of feature strength)  and difficulty in precise $\text{L}0$ control \citep{gao2024scaling, templeton2024scaling} have driven architectural innovation. JumpReLU SAEs \citep{rajamanoharan2024jumping}, use a learned threshold within the activation function for direct $\text{L}0$ optimization. TopK SAE \citep{gao2024scaling} selects only the top $K$ activations, simplifying tuning and reducing shrinkage compared to L1; BatchTopK SAE \citep{bussmann2024batchtopk} further improves it by applying the TopK constraint at the batch level for adaptive sparsity and improved reconstruction.  The latter approach appears promising for our purposes, as it allows precise sparsity control along with excellent reconstruction capabilities. For a detailed overview of the SAE variations, we further refer the reader to the survey by \cite{shu2025survey}.

Despite ongoing advancements, SAEs continue to face challenges:  \citet{chanin2024absorption} describes the phenomenon of feature \textit{absorption}, when broad features absorb into more specific ones. \citet{leask2025sparse} highlights feature \textit{composition}, when independent features merge into one larger feature, and introduces the MetaSAE, which emerges as a promising approach to identify these problems.

Recently, \citet{bussmann2025learning} proposed Matryoshka SAE to address these issues by employing a hierarchical approach. It builds upon BatchTopK architecture and uses nested features with increasing latent space size so that SAE separately learns broad and specific features. However, this hierarchical design leads to feature hedging \citep{chanin2025feature}, where narrow higher-level dictionaries merge correlated features, reducing interpretability. Additionally, this approach introduces substantial computational overhead ($+50\%$ compared to traditional SAEs) and a degradation in reconstruction performance. Furthermore, while its reliance on hierarchical representation seems intuitive, its interpretability remains poorly explored. In contrast, we explore an alternative direction of representations decorrelation \citep{cogswell2016reducingoverfittingdeepnetworks, wang2021mmaregularizationdecorrelatingweights, rodríguez2017regularizingcnnslocallyconstrained}. OrtSAE proposes an efficient approach by directly enforcing the orthogonality between SAE latents. It avoids all of the mentioned problems while achieving performance similar to Matryoshka SAEs in mitigating the feature absorption and composition challenges.

\section{Orthogonal Sparse Autoencoders}
\label{sec:Orthogonal SAE}

\subsection{Traditional Sparse Autoencoders}
\label{subsec:traditional_sae}

Sparse autoencoders (SAEs) aim to reconstruct model activations \(\mathbf{x} \in \mathbb{R}^n\) as a sparse linear combination of $m \gg n$ feature vectors, or \textit{latents}. Formally, a SAE consists of an encoder and a decoder:
\begin{equation}
\begin{aligned}
  \mathbf{h(x)} &= \sigma(\mathbf{W}^{\text{enc}} \mathbf{x} + \mathbf{b}^{\text{enc}}), \\
  \hat{\mathbf{x}}(\mathbf{h}) &= \mathbf{W}^{\text{dec}} \mathbf{h} + \mathbf{b}^{\text{dec}}.
\end{aligned}
\end{equation}

The encoder, followed by a non-linearity $\sigma(\cdot)$, learns a mapping from the activations to a sparse and overcomplete latent code $\mathbf{h(x)} \in R^m$. Given $\mathbf{h(x)}$, the decoder reconstructs the original input as a sparse linear combination of latents, $\mathbf{W^{\text{dec}}_i}$, $\mathbf{i} = 1,...,m$, as $\hat{\mathbf{x}}$.

The standard loss function to train a SAE is defined as:
\begin{equation}
\label{eq:sae_loss}
  \mathcal{L}(\mathbf{x}) = \underbrace{\left\|\mathbf{x} - \hat{\mathbf{x}}(\mathbf{h(x)})\right\|_2^2}_{\mathcal{L}_\text{reconstruct}} + \underbrace{\lambda\,S(\mathbf{h(x)})}_{\mathcal{L}_\text{sparsity}} + \alpha L_{\text{aux}},
\end{equation}
where $S$ is a sparsity penalty and $\lambda$ is a coefficient controlling the trade-off between sparsity and reconstruction quality. The optional \(L_{\text{aux}}\) term covers any auxiliary penalties (e.g. recycling dead units \citep{gao2024scaling}).

Traditional SAEs focus on reducing reconstruction loss while increasing sparsity. The ReLU SAE \citep{bricken2023towards, cunningham2023sparse} uses the ReLU activation function and applies an $\text{L}1$ penalty to ensure sparsity in $\mathbf{h(x)}$. TopK SAE \citep{gao2024scaling} achieves sparsity by zeroing all entries of $\mathbf{h(x)}$ except for the $K$ largest ones. BatchTopK \citep{bussmann2024batchtopk} SAE further improves the idea by selecting the top $B \times K$ entries across a batch of $\mathbf{h(x)}$, allowing some examples to have more or less active latents.

\subsection{Challenges in Training SAEs}
\label{subsec:orthogonal_intuition}

Traditional SAEs training objectives (Eq. \ref{eq:sae_loss}) combine reconstruction loss and sparsity penalty. While sparsity is required to decompose activations into interpretable features, optimization of it results in multiple failure modes. \textit{Feature absorption} (Fig. \ref{fig:absorption}) occurs when an interpretable feature becomes SAE latent which appears to represent that feature, but fails to fire on arbitrary tokens that it seemingly should activate on. Instead, token-aligned latents fire, ``absorbing'' part of the feature representation to satisfy the sparsity objective by activating fewer latents overall. \textit{Feature composition} (Fig. \ref{fig:composition}) occurs when features overlap. To optimize over sparsity, SAE learns a single latent that captures the specific combination of features (e.g. ``red square'') instead of representing the underlying features (``red'' and ``square'') with separate latents.

Feature absorption and composition produce redundant representations where multiple latents capture overlapping concepts, leading to high cosine similarities between decoder vectors. Formally, consider two atomic features A (``red'') and B (``square'') (Fig. \ref{fig:composition}). In traditional SAEs, these independent features can merge into a composite feature C (``red square''). Let $\mathbf{W}_A^{\text{dec}}, \mathbf{W}_B^{\text{dec}}$, and $\mathbf{W}_C^{\text{dec}}$ denote the decoder vectors for features A, B, and C, respectively. When feature composition occurs, C incorporates components of both features A and B. This creates higher correlations between C and each atomic feature: $\cos(\mathbf{W}_C^{\text{dec}}, \mathbf{W}_A^{\text{dec}}) > \cos(\mathbf{W}_A^{\text{dec}}, \mathbf{W}_B^{\text{dec}})$ and $\cos(\mathbf{W}_C^{\text{dec}}, \mathbf{W}_B^{\text{dec}}) > \cos(\mathbf{W}_A^{\text{dec}}, \mathbf{W}_B^{\text{dec}})$. Similarly, feature absorption creates overlapping latents, resulting in the decoder vectors that are more correlated than they should be for truly atomic features. To address these issues, OrtSAE extends the traditional SAE objective by enforcing orthogonality between SAE latents. At each training step, we penalize high cosine similarities between SAE latents, directly approaching both absorption and composition problems by encouraging the formation of more atomic features.

% Feature absorption and composition produce redundant representations in which multiple latents capture overlapping features, resulting in high cosine similarities between them. OrtSAE extends the traditional SAE objective by enforcing orthogonality between SAE latents. At each training step, we penalize high cosine similarities between SAE latents, directly approaching both absorption and composition problems by encouraging the formation of more atomic features.

\subsection{OrtSAE Training Procedure}
\label{subsec:ortsae_training}

The main contribution of OrtSAE is the introduction of a new orthogonalization penalty that penalizes high similarities between SAE latents. Formally, given a SAE with decoder matrix $\mathbf{W^{\text{dec}}} \in \mathbb{R}^{n \times m}$, we first define the cosine similarity between two feature vectors as:
\begin{equation}
\cos(\mathbf{W^{\text{dec}}_i}, \mathbf{W^{\text{dec}}_j}) = \frac{\langle \mathbf{W^{\text{dec}}_i}, \mathbf{W^{\text{dec}}_j} \rangle}{\max(\|\mathbf{W^{\text{dec}}_i}\|_2 \cdot \|\mathbf{W^{\text{dec}}_j}\|_2, \delta)},
\end{equation}
where $\delta > 0$ is a small constant added to prevent division by zero. Using this definition, we formulate our orthogonality penalty as:
\begin{equation}
L_{\text{orthogonal}}(\mathbf{W}^{\text{dec}}) = \frac{1}{K(m)}\sum_{k=1}^{K(m)}\frac{1}{|C_k|}\sum_{i \in C_k}\left(\max_{\substack{j \in C_k \\ j \neq i}}\cos(\mathbf{W^{\text{dec}}_i}, \mathbf{W^{\text{dec}}_j})\right)^2.
\end{equation}
Instead of computing all pairwise similarities between feature vectors $\mathbf{W^{\text{dec}}_i}$, which would require $\mathcal{O}(m^2)$ operations and is infeasible for large $m$, at each training step we \textit{randomly} partition the latent space into $K := K(m)$ equal chunks, each containing a fixed number of latents, $|C_k|, k=1,...,K(m)$, proportional to $m$. Within each $k$-th chunk, we find the maximum pairwise cosine similarity between every $\mathbf{W^{\text{dec}}_i}$, $\mathbf{i} \in C_k$ and all other latents from $C_k$, square this value to penalize highly correlated features more and compute the expectation. We compute the final value by averaging across all $K$ chunks. This chunk-wise strategy reduces the computational complexity to $\mathcal{O}(m)$ and provides an efficient scaling strategy to a larger latent spaces.

With the orthogonal penalty defined, OrtSAE training objective is defined as:
\begin{equation}
\mathcal{L}_{\text{OrtSAE}}(x) = L_{\text{MSE}} + \lambda L_{\text{sparsity}} + \alpha L_{\text{aux}} + \gamma L_{\text{orthogonal}},    
\end{equation}
where $\gamma$ is an \textit{orthogonality coefficient} that controls the strength of the applied penalty. 

To further enhance computational efficiency, we explore computing the orthogonality loss every fifth training iteration, scaling the orthogonality coefficient $\gamma$ by a factor of 5 to maintain regularization strength, which yields comparable performance with significantly reduced computational overhead, as detailed in Appendix~\ref{app:chunk_analysis}.

% \begin{tcolorbox}[
%     enhanced,
%     breakable,
%     colback=gray!10,      % light gray background
%     colframe=gray!50,     % darker gray frame
%     boxrule=0.5pt,        % frame thickness
%     arc=2pt,              % rounded corners
%     left=4pt, right=4pt,  % horizontal padding
%     top=4pt, bottom=4pt   % vertical padding
% ]
% \begin{verbatim}
% def orthogonal_loss(W_dec, num_chunks=10):
%     """
%     W_dec: [d_model, n_features] decoder matrix
%     num_chunks: number of partitions for scaling
%     """
%     loss = 0
%     features = random_permutation(W_dec.columns)
%     chunks = split_into_random_chunks(features, num_chunks)
    
%     for chunk in chunks:
%         # Compute pairwise cosine similarities
%         W_chunk = W_dec[:, chunk]
%         gram_matrix = W_chunk.T @ W_chunk  
        
%         # For each feature, find max similarity to others
%         for i in range(len(chunk)):
%             others = [j for j in range(len(chunk)) if j != i]
%             max_sim = max(gram_matrix[i, others])
%             loss += max_sim**2
            
%     return loss
% \end{verbatim}
% \end{tcolorbox}

\section{Experiments}
\label{sec:Experiments}

Here, we present our experimental evaluation of OrtSAE. Sec.~\ref{subsec:Experimental Setup} covers the experimental setup. Sec.~\ref{subsec:Fundamental Performance Metrics} compares OrtSAE’s core performance metrics to other methods. Sec.~\ref{subsec:Atomicity investigation} presents quantitative and qualitative results on feature atomicity. Sec.~\ref{subsec:Downstream benchmarks} assesses OrtSAE’s performance on downstream tasks using SAEBench.

% We evaluate OrtSAE against BatchTopK SAE and ReLU SAE baselines through three central research questions:

% \begin{enumerate}
% \item \textbf{Does orthogonal regularization preserve SAE reconstruction performance?} We compare core metrics including explained variance and KL-divergence to assess fidelity.
    
% \item \textbf{Does orthogonal regularization improve feature atomicity?} We analyze feature absorption/composition rates and feature clustering patterns.
    
% \item \textbf{Does orthogonal regularization enhance downstream task performance?} We evaluate using SAEBench's spurious correlation removal and targeted perturbation tasks, sparse probing and RAVEL.
% \end{enumerate}

\subsection{Experimental Setup.}
\label{subsec:Experimental Setup}
\paragraph{Baselines.} 
We compare our model with: 1) traditional SAEs, such as ReLU SAE \citep{bricken2023towards, cunningham2023sparse} and state-of-the-art BatchTopK SAE \citep{bussmann2024batchtopk}; 2) recent Matryoshka SAE that enforce nested, hierarchical learning at multiple feature levels \citep{bussmann2025learning}.

\paragraph{Models Configuration.} 
Following the work of \cite{bussmann2025learning} we train SAEs on the activations from layer 12 of the Gemma-2-2B (26 layers total). Each SAE has latent space of size $m = 65536$ and sparsity levels $\text{L}0$ in $\{25, 40, 55, 70, 85, 100, 115, 130\}$. The training uses 500 million tokens from the OpenWebText dataset \citep{gokaslan2019} with a context length of 1024. 
As a basis of OrtSAE, we follow BatchTopK SAE \href{https://github.com/saprmarks/dictionary_learning.git}{repository}, leveraging BatchTopK’s precise $\text{L}0$ sparsity control.
For OrtSAE, we set the number of chunks $K(m) = \lceil m / 8192 \rceil$ (yielding $K = 8$ for $m = 65536$) with $\gamma = 0.25$.
The full details and hyperparameters are available in the Appendix \ref{app:sae_details}.  To assess the transferability of our approach, we also conduct experiments on layer 20 of Gemma-2-2B and layer 20 of Llama-3-8B and report the results in Appendix \ref{app:additional_experiments}.To ensure the reproducibility, we will publicly release all code and hyperparameters.

\paragraph{Evaluation.} 
Following \cite{gao2024scaling} and \cite{  bussmann2025learning}, we evaluate SAEs core performance through explained variance, KL-divergence, orthogonality, and feature interpretability. Atomicity analysis includes absorption metrics, MetaSAE-based composition rate \citep{leask2025sparse}, clustering, cross-model overlap, and qualitative investigation. Downstream assessment uses SAEBench \citep{karvonen2025saebench} with spurious correlation removal, targeted probe perturbation, sparse probing, and RAVEL tasks.

\subsection{Foundational Performance Analysis}
\label{subsec:Fundamental Performance Metrics}

\begin{figure}[t!]
    \centering
    \begin{subfigure}[t]{0.245\textwidth}
        \centering
        \includegraphics[width=\textwidth]{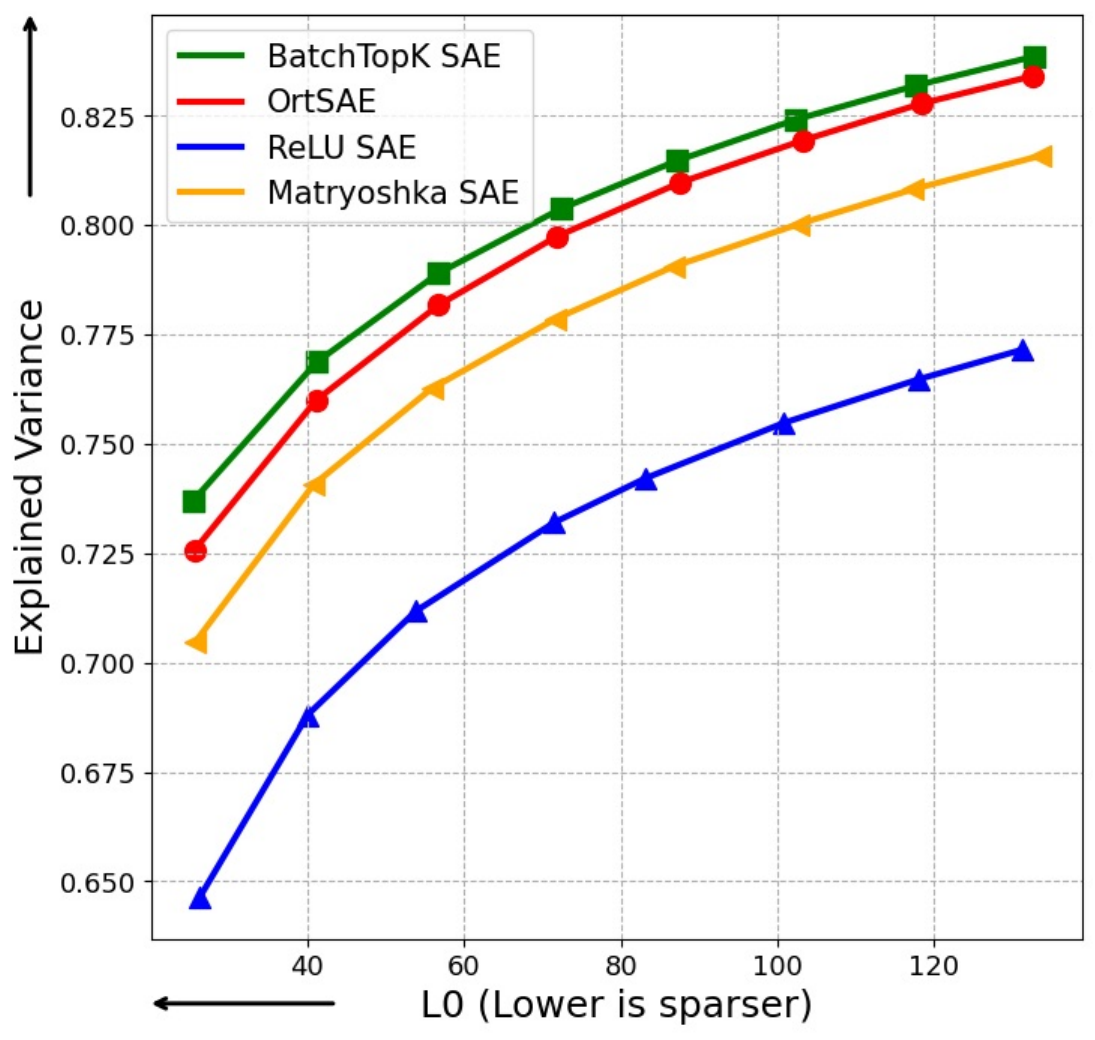}
        \caption{Explained variance.}
        \label{fig:explained_variance}
    \end{subfigure}
    \hfill
    \begin{subfigure}[t]{0.245\textwidth}
        \centering
        \includegraphics[width=\textwidth]{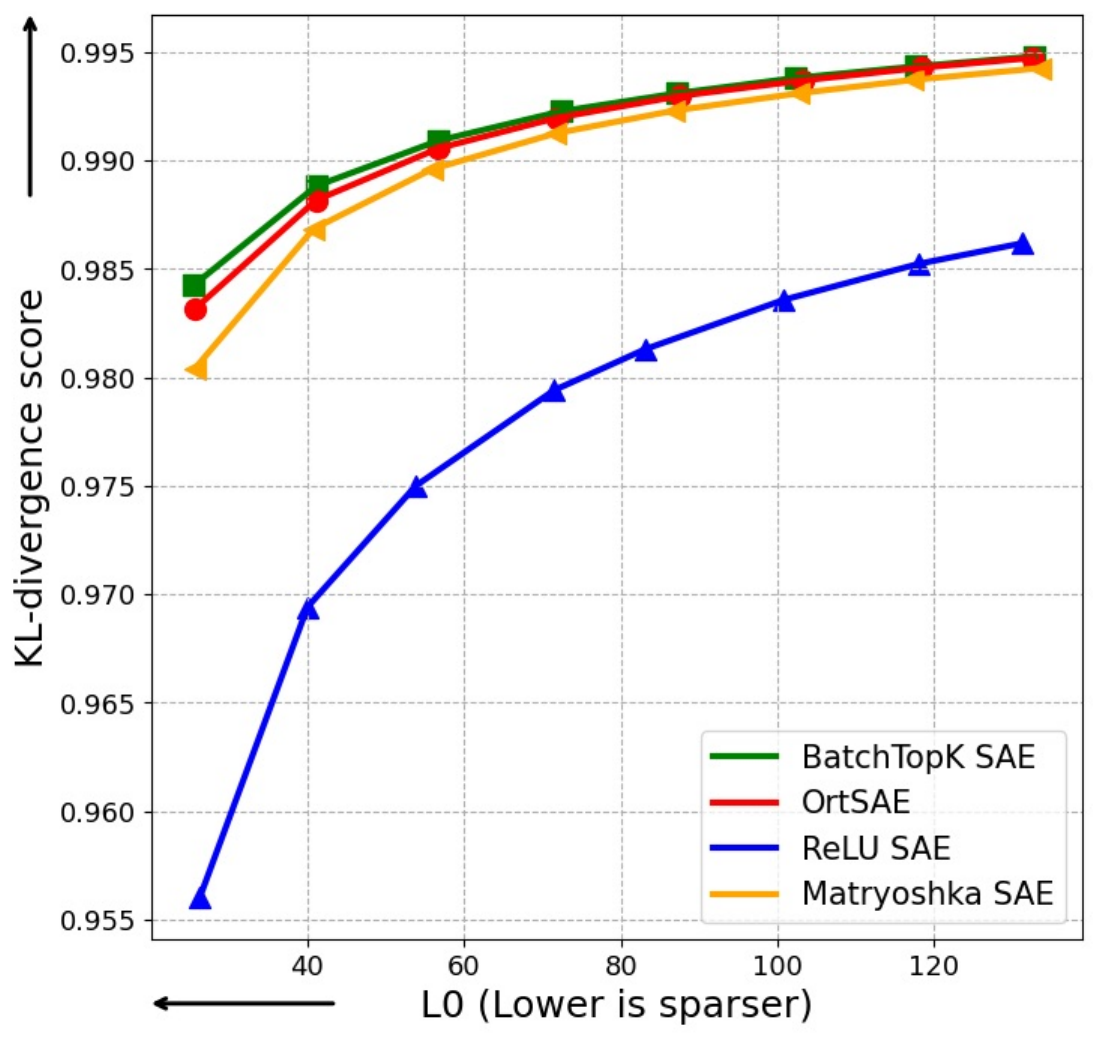}
        \caption{KL divergence score.}
        \label{fig:kl_div_score}
    \end{subfigure}
    \hfill
        \begin{subfigure}[t]{0.24\textwidth}
        \centering
        \includegraphics[width=\textwidth]{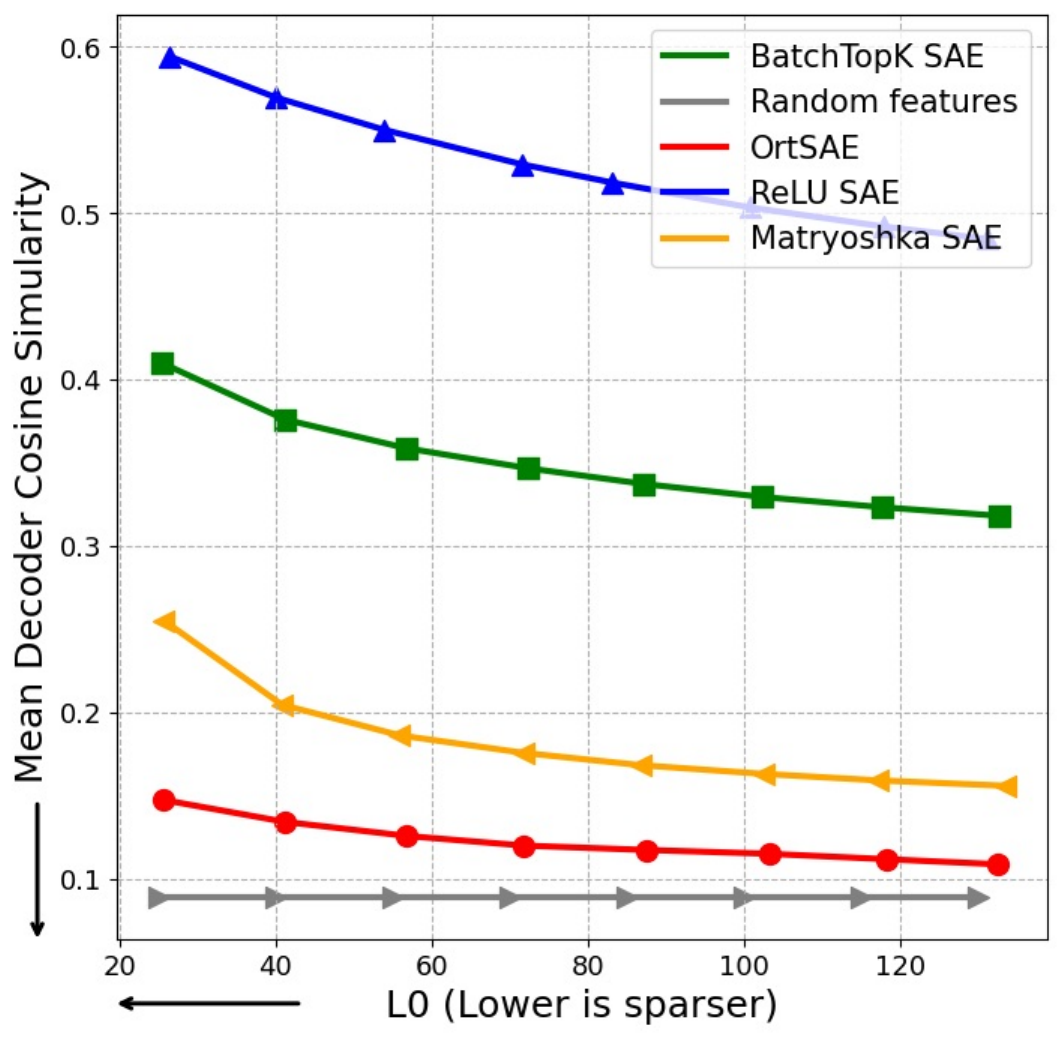}
        \caption{Mean cosine similarity.}
        \label{fig:orthogonality}
    \end{subfigure}
    \hfill
    \begin{subfigure}[t]{0.245\textwidth}
        \centering
        \includegraphics[width=\textwidth]{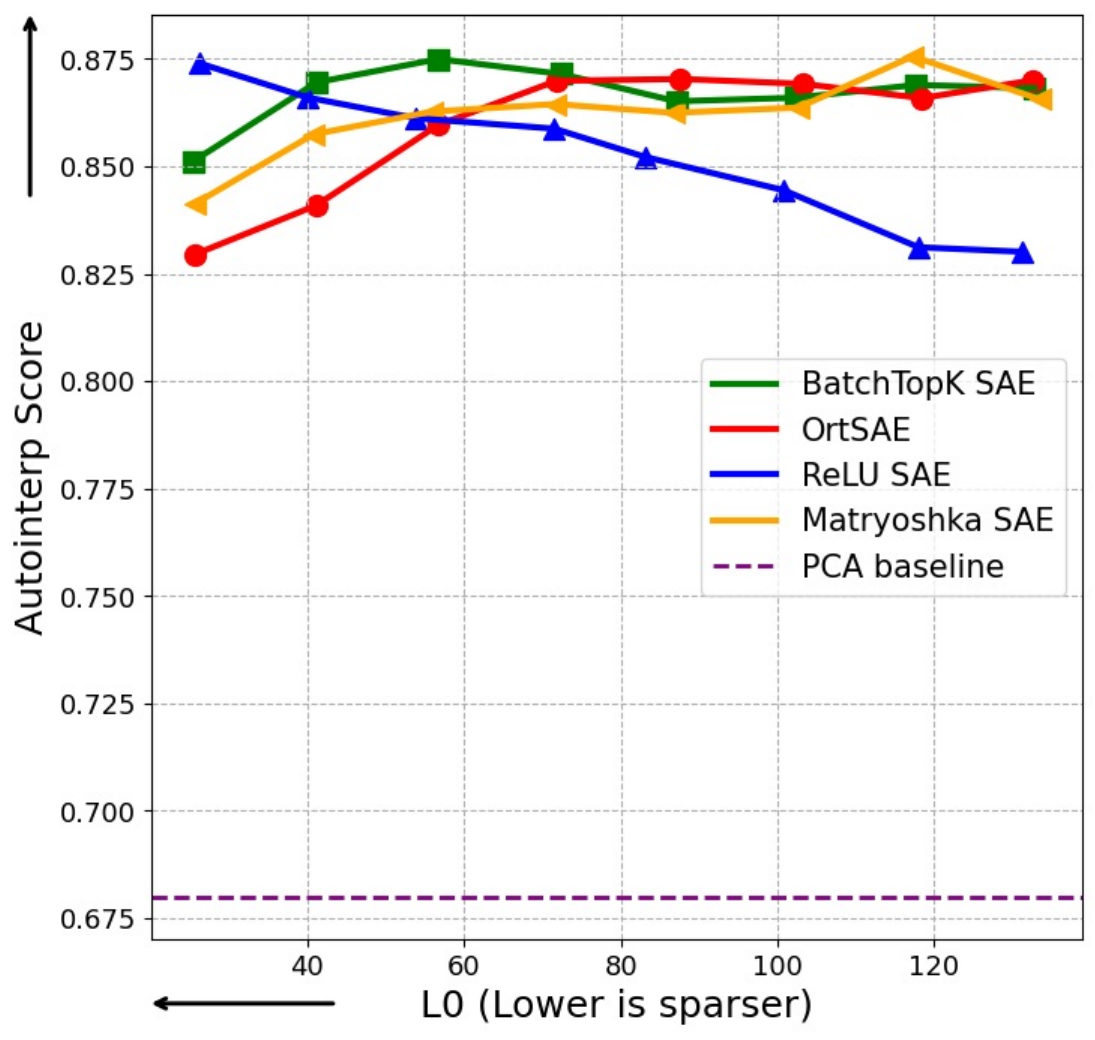}
        \caption{Autointerp Score.}
        \label{fig:Autointerp Score}
    \end{subfigure}
    \caption{\textbf{Core performance metrics.} (a) Explained variance: OrtSAE shows slightly lower reconstruction fidelity than BatchTopK SAE but outperforms Matryoshka SAE. (b) KL-divergence score: OrtSAE matches BatchTopK SAE and exceeds Matryoshka SAE. (c) Mean cosine similarity to closest decoder feature: OrtSAE achieves near-random initialization orthogonality, significantly lower than other SAE variants. (d) Autointerp Score: OrtSAE demonstrates interpretability comparable to both BatchTopK and Matryoshka SAEs.}
    \label{fig:core_metrics}
\end{figure}

We evaluate OrtSAE using four metrics: (1) reconstruction fidelity, computed as the fraction of explained variance; (2) downstream predictive performance, measured by the KL-divergence score between the original LLM's output distributions and those generated using reconstructed activations; (3) feature orthogonality, calculated as the mean cosine similarity to each feature's nearest decoder neighbor; and (4) feature interpretability, assessed via the Autointerp Score. Fig.~\ref{fig:core_metrics} shows these metrics across sparsity levels, demonstrating OrtSAE's balance between reconstruction quality and model functionality preservation. To further validate the generalizability of our findings across different model architectures and layers, we conducted additional experiments on layer 20 of Gemma-2-2B and layer 20 of Llama-3-8B, with results reported in App.~\ref{app:additional_experiments}. We also analyze the impact of varying the number of chunks on OrtSAE performance to assess its robustness and scalability, with details provided in App.~\ref{app:chunk_analysis}.

% We evaluate the core performance of OrtSAE against baselines by analyzing four key metrics: reconstruction fidelity (via explained variance), downstream predictive performance (via KL-divergence score), feature orthogonality (via mean cosine similarity), and interpretability (via Autointerp Score). These metrics, shown in Fig.~\ref{fig:core_metrics}, collectively assess how well OrtSAE balances activation reconstruction, functional preservation, feature separation, and human-understandable representations across various sparsity levels (L0).

% \textbf{Reconstruction and Predictive Performance.} Reconstruction fidelity, quantified by explained variance, measures how accurately the SAE reconstructs input activations. Downstratream predictive performance, assessed via KL-divergence score, evaluates preservation of the language model’s predictive behavior using reconstructed activations. Fig.~\ref{fig:core_metrics}a shows BatchTopK SAE slightly outperforms OrtSAE in explained variance, indicating a minor trade-off due to orthogonality constraints. However, OrtSAE retains sufficient reconstruction quality. Fig.~\ref{fig:core_metrics}b demonstrates OrtSAE’s near-identical KL-divergence scores to BatchTopK SAE across sparsity levels. This indicates that OrtSAE discards reconstruction noise (data minimizing MSE but irrelevant to predictions), preserving critical functional representations.

\paragraph{Reconstruction and Predictive Performance.} We first evaluate reconstruction quality through explained variance, measuring how accurately each SAE reconstructs input activations. To assess whether these reconstructions preserve the model's functionality, we additionally examine downstream predictive performance using KL-divergence scores (detailed descriptions provided in App.\ref{app:KL-divergence score definition}). We also tested the effect of decoded activations on the base language model’s perplexity using LogLoss, which shows the same patterns as the KL-divergence scores (see App.~\ref{app:logloss}). Fig.~\ref{fig:core_metrics}a shows OrtSAE achieves comparable performance to BatchTopK SAE while outperforming Matryoshka SAE by $2\%$ in fraction of explained variance. The KL-divergence scores (Fig.~\ref{fig:core_metrics}b) reveal nearly identical predictive behavior between OrtSAE and BatchTopK SAE, with both slightly surpassing Matryoshka SAE. These results are particularly notable because OrtSAE maintains strong performance despite its additional orthogonality constraints, whose effects we analyze next through feature similarity.

\paragraph{Feature Orthogonality.} To quantify the separation of decoder features, we compute the mean cosine similarity (MeanCosSim) for each feature vector \(i\) to its closest neighbor in the decoder matrix:
\begin{equation}
\text{MeanCosSim} = \frac{1}{m}\sum_{i=1}^m \max_{j \neq i} \cos(\mathbf{W}^{\text{dec}}_i, \mathbf{W}^{\text{dec}}_j).
\end{equation}
As shown in Fig.~\ref{fig:core_metrics}c, OrtSAE achieves superior separation with MeanCosSim values 2.7 times lower than BatchTopK and 1.5 times lower than Matryoshka SAE, approaching random initialization levels. This enhanced orthogonality directly contributes to improved feature atomicity, as shown in Sec.~\ref{subsec:Atomicity investigation}.

\paragraph{Feature Interpretability.} The Autointerp Score evaluates feature interpretability using GPT-4o-mini as an LLM judge. Following \cite{paulo2024automatically}'s methodology, we first generate interpretable descriptions of 1,000 latents using an LLM, then quantitatively assess these explanations by measuring how accurately the LLM can predict whether each latent activates (fires) on new input tokens (detailed descriptions provided in Appx. \ref{app:autointerp}). Fig.~\ref{fig:core_metrics}d demonstrates comparable interpretability between OrtSAE, BatchTopK SAE, and Matryoshka SAE latents across all tested sparsity levels.

These results demonstrate that OrtSAE achieves an optimal balance - matching top reconstruction performance while significantly improving feature separation without compromising interpretability. This demonstrates that our new training procedure can enhance feature quality while preserving model functionality, as we explore further in our atomicity analysis.

\subsection{Atomicity Analysis}
\label{subsec:Atomicity investigation}

\begin{figure}[t!]
    \centering
    \begin{subfigure}[t]{0.25\textwidth}
        \centering
        \includegraphics[width=\textwidth]{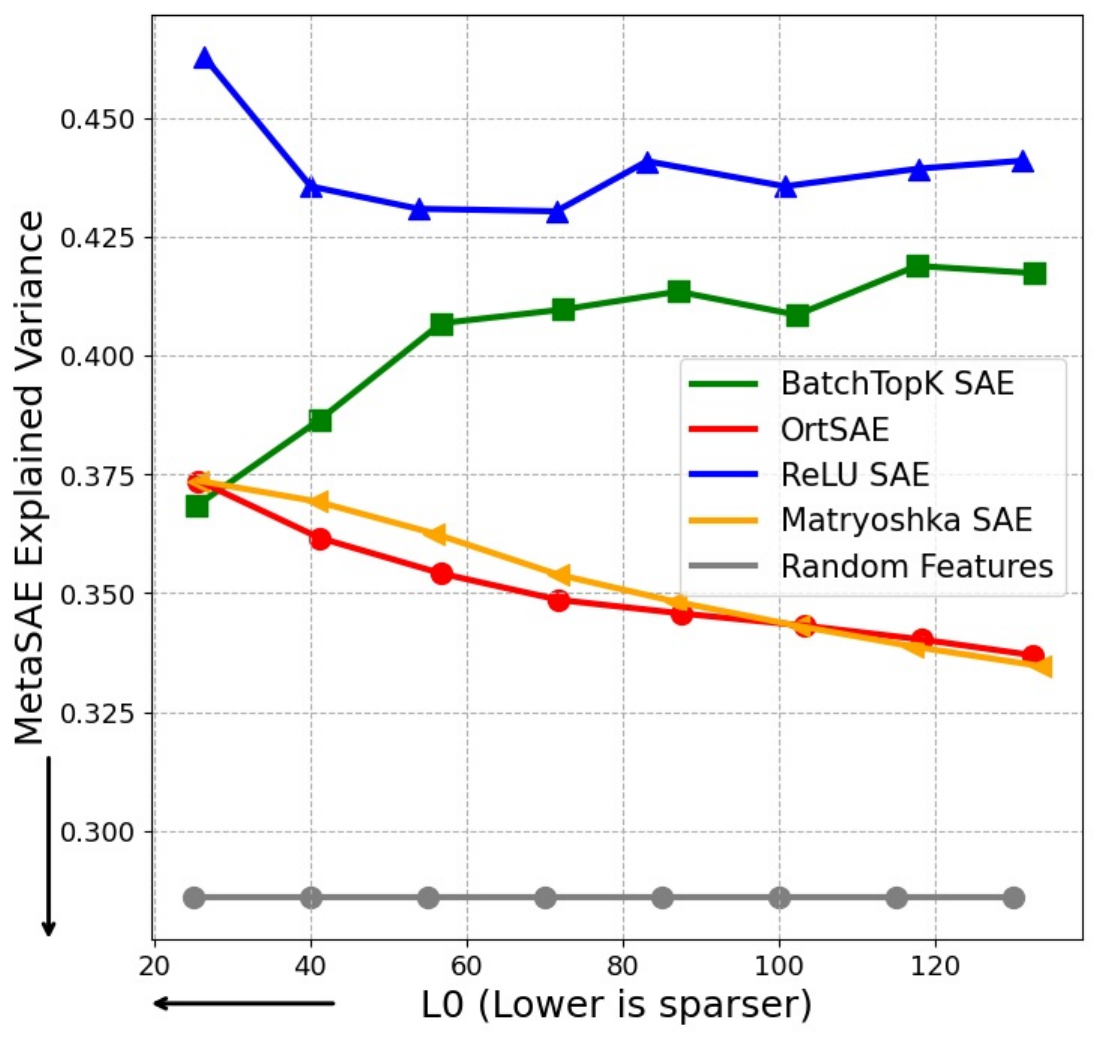}
        \caption{Composition rate.}
        \label{fig:explained_variance}
    \end{subfigure}
    \begin{subfigure}[t]{0.243\textwidth}
        \centering
        \includegraphics[width=\textwidth]
        {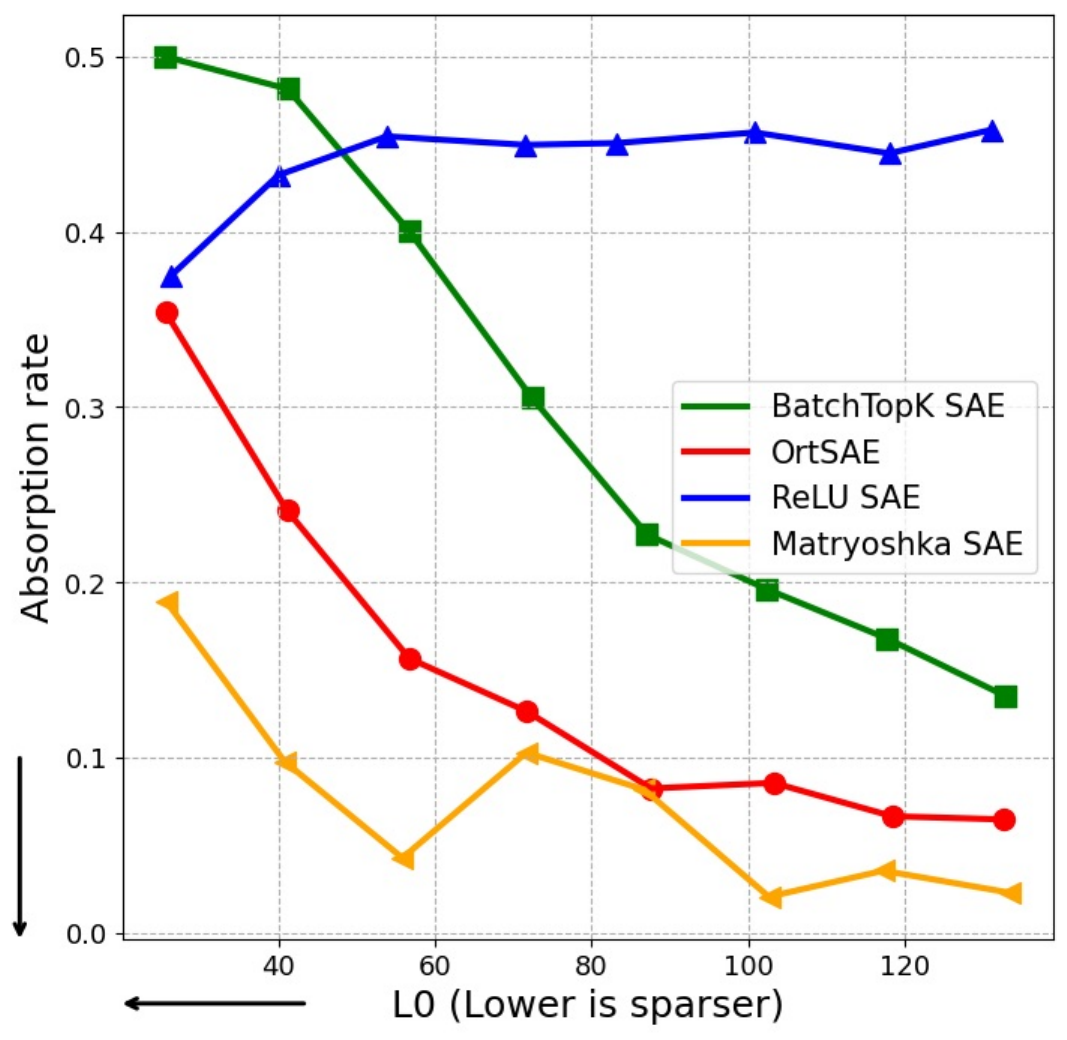}
        \caption{Absorption rate.}
        \label{fig:ce_loss_score}
    \end{subfigure}
    \begin{subfigure}[t]{0.243\textwidth}
        \centering
        \includegraphics[width=\textwidth]
        {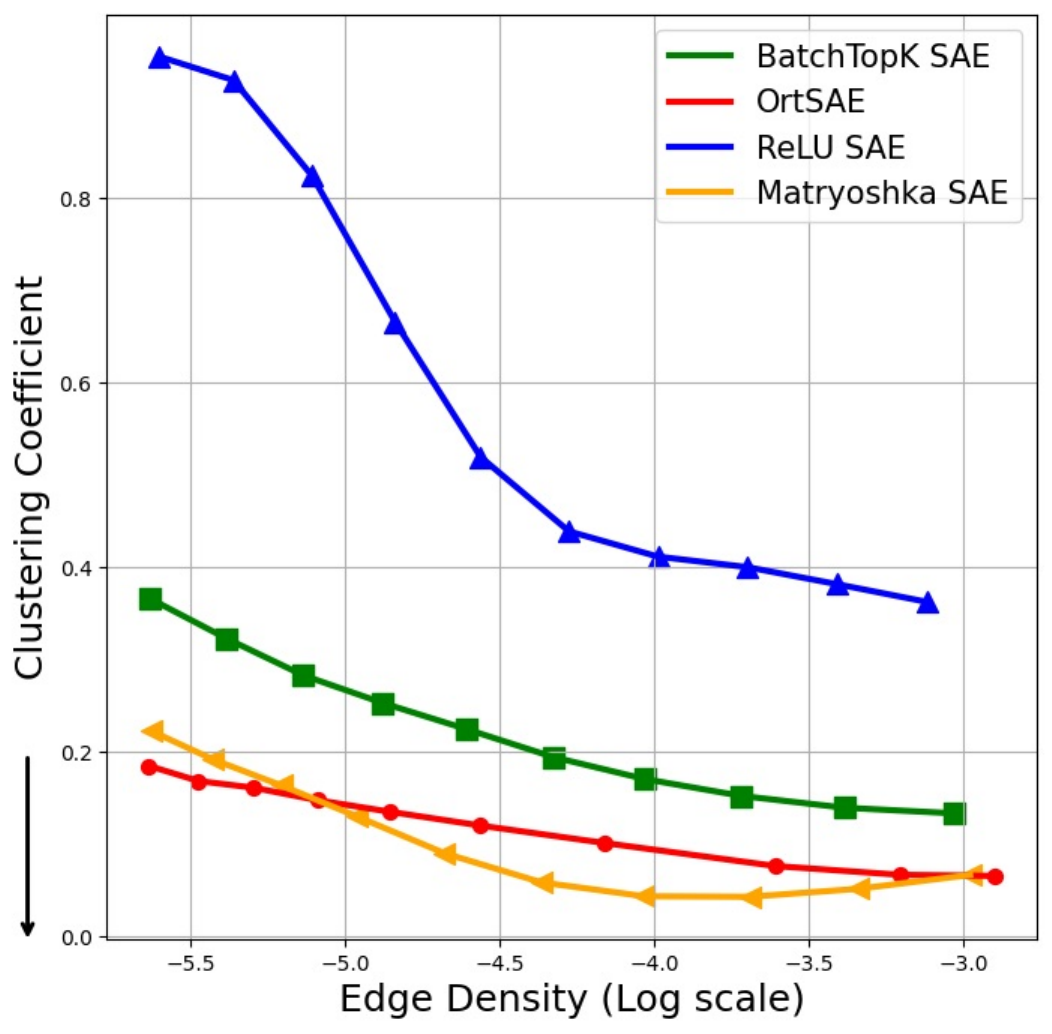}
        \caption{Clustering rate.}
        \label{fig:ce_loss_score}
    \end{subfigure}
    \begin{subfigure}[t]{0.243\textwidth}
        \centering
        \includegraphics[width=\textwidth]
        {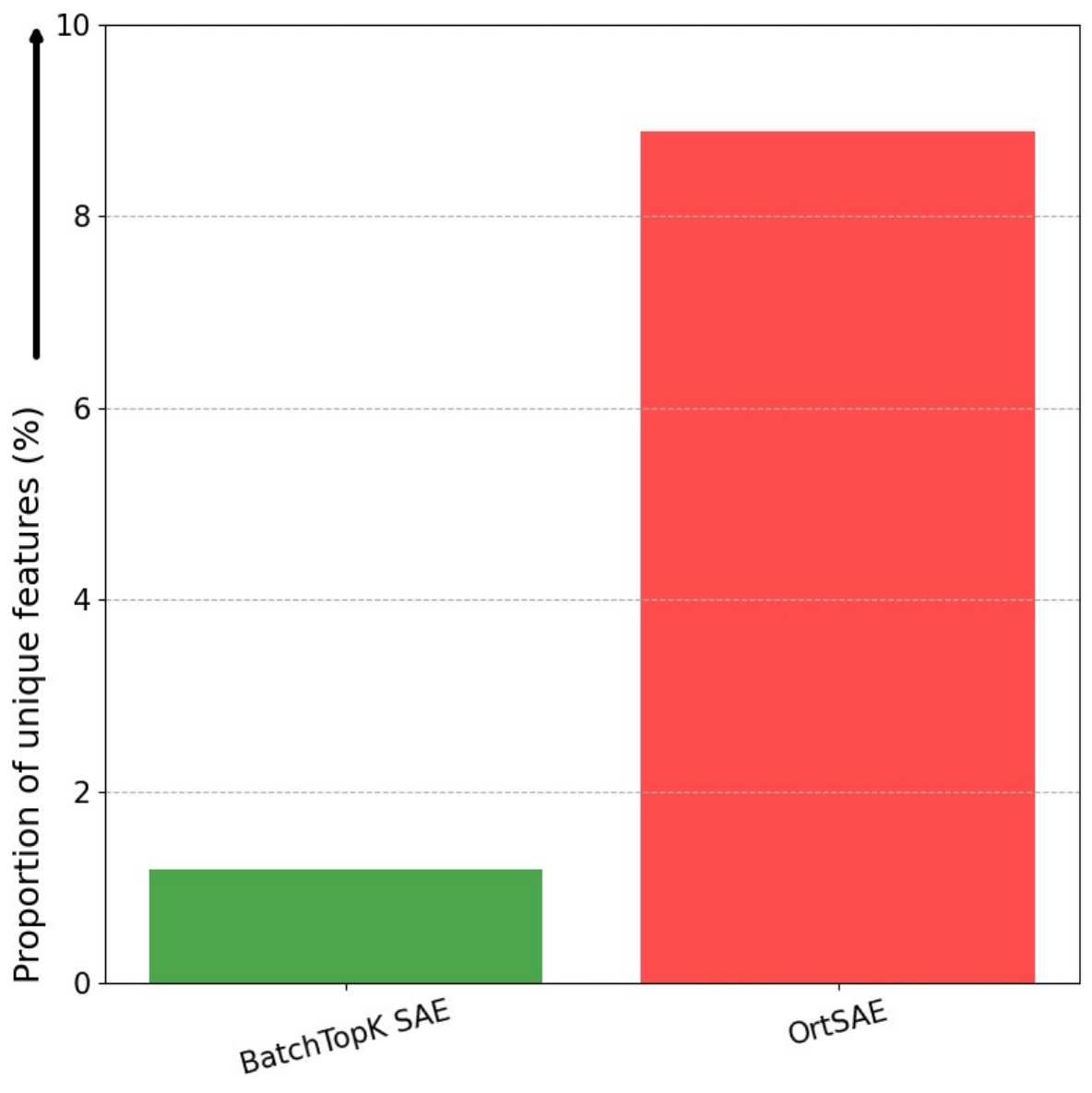}
        \caption{Unique features.}
        \label{fig:ce_loss_score}
    \end{subfigure}
    \caption{\textbf{Atomicity metrics.} (a) MetaSAE-based Composition rate: OrtSAE shows reduced feature merging compared to traditional SAEs and matches Matryoshka SAE. (b) Absorption rate: OrtSAE shows significant improvement over BatchTopK with performance approaching Matryoshka SAE. (c) Feature clustering: OrtSAE matches Matryoshka SAE in low feature interconnection, both outperforming traditional SAEs. (d) Proportion of unique features: OrtSAE discovers substantially more distinct features than BatchTopK SAE.}
    \label{fig:atomicity_metrics}
\end{figure}

\begin{figure}[t!]
    \centering
    \begin{subfigure}[t]{0.32\textwidth}
        \centering
        \includegraphics[width=\textwidth]{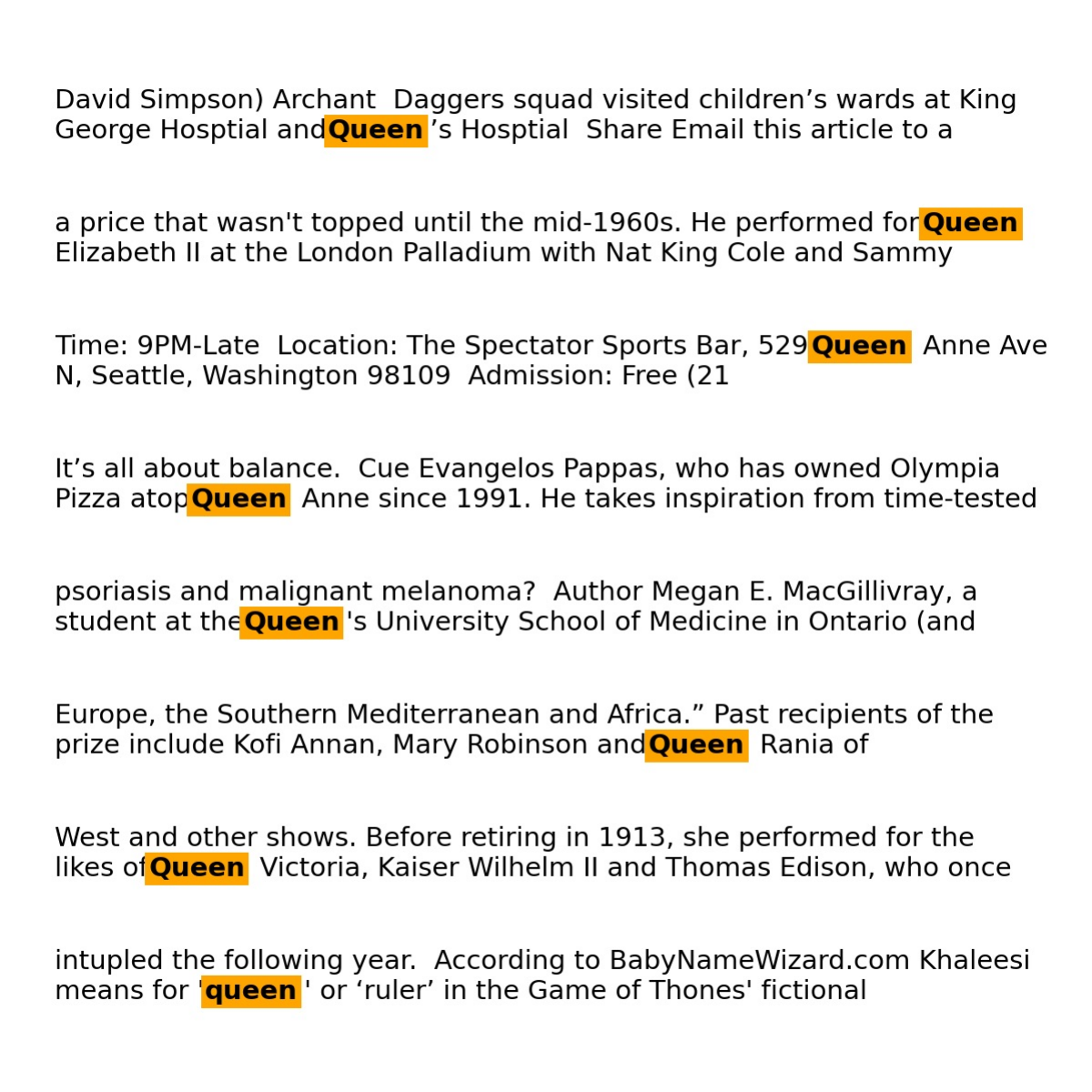}
        \caption{BatchTopK SAE feature that \\ activates only on ``Queen'' token}
        \label{fig:explained_variance}
    \end{subfigure}
    \begin{subfigure}[t]{0.32\textwidth}
        \centering
        \includegraphics[width=\textwidth]
        {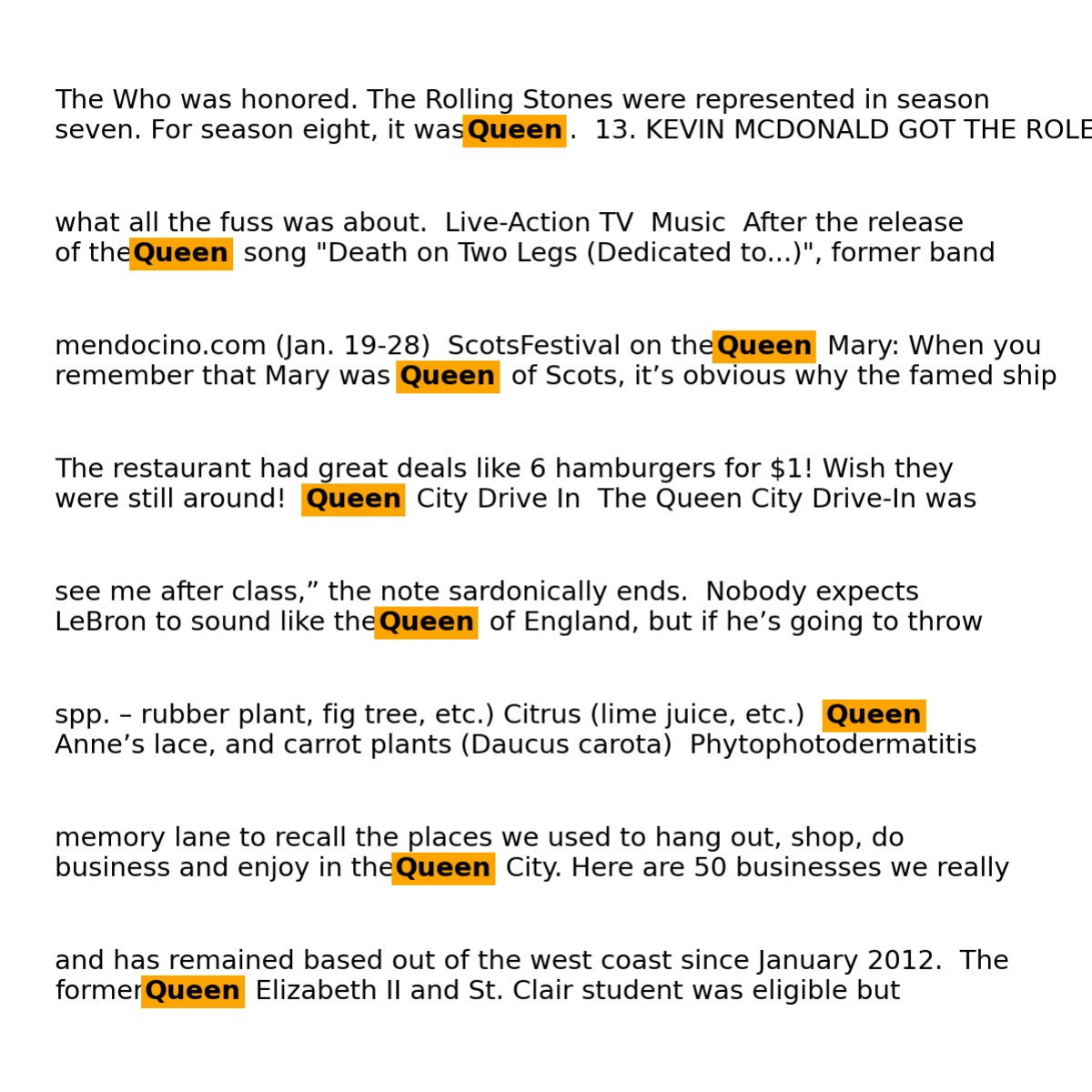}
        \caption{OrtSAE feature that activates only on ``Queen'' token}
        \label{fig:ce_loss_score}
    \end{subfigure}
    \begin{subfigure}[t]{0.32\textwidth}
        \centering
        \includegraphics[width=\textwidth]
        {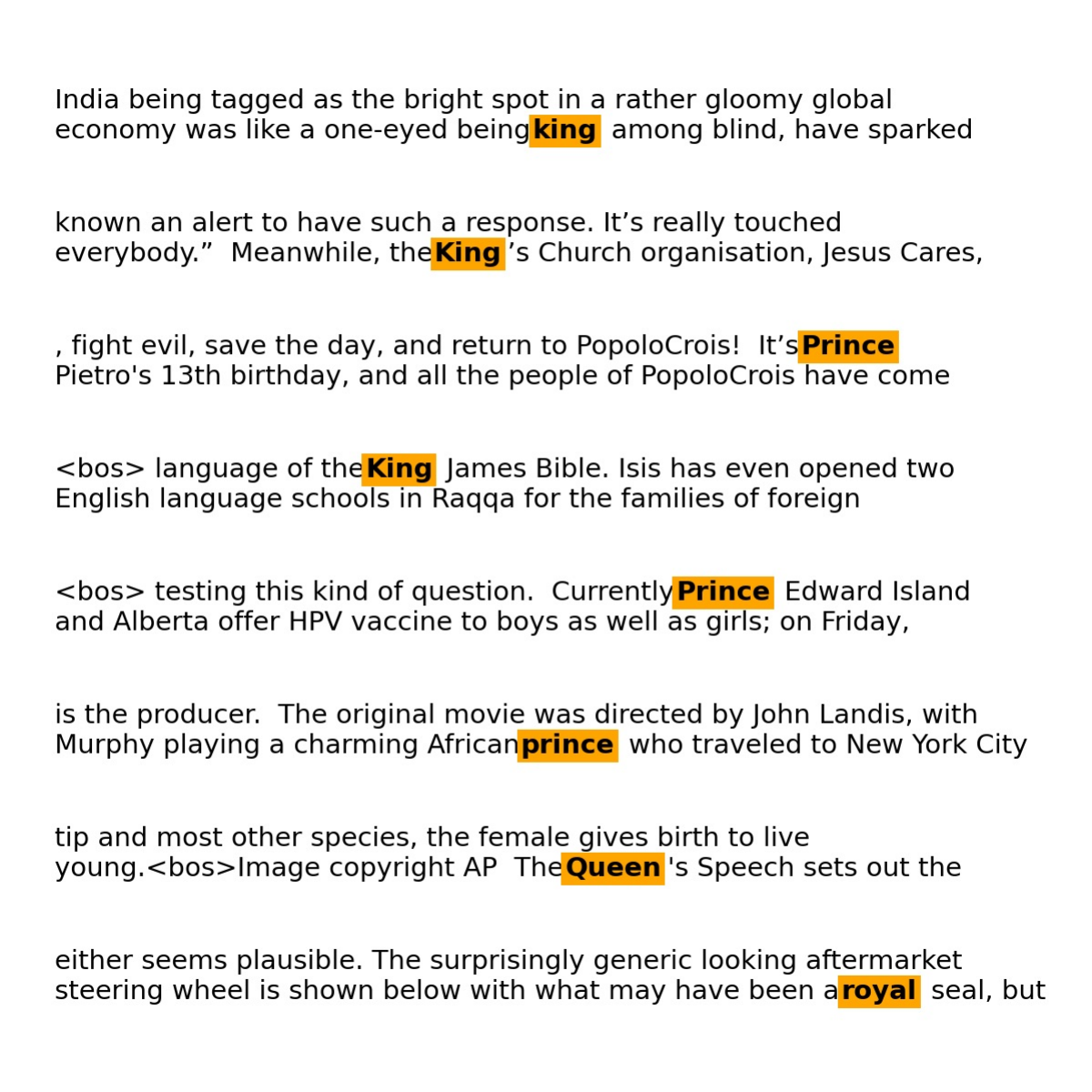}
        \caption{OrtSAE feature that activates only on titles and royal concepts}
        \label{fig:ce_loss_score}
    \end{subfigure}
    \caption{\textbf{Decomposition of a BatchTopK SAE Feature into OrtSAE Features.} (a) A BatchTopK SAE feature activating solely on the token ``Queen'', expressed as a linear combination of two OrtSAE features: (b) An OrtSAE feature specific to the token ``Queen''. (c) An OrtSAE feature capturing royal titles, revealing how OrtSAE disentangles the broader royalty concept absorbed by the ``Queen'' feature in BatchTopK SAE.}
    \label{fig:queen_example}
\end{figure}

We assess feature atomicity through four quantitative measures: (1) MetaSAE-based composition rates, (2) absorption metrics, (3) clustering coefficients, and (4) cross-model feature uniqueness, complemented by qualitative analysis of decomposed features. Fig.~\ref{fig:atomicity_metrics} presents the quantitative comparisons across sparsity levels, while Fig.~\ref{fig:queen_example} illustrates OrtSAE's ability to disentangle composite features through concrete examples.

\paragraph{MetaSAE-Based Feature Composition Analysis.} We train a MetaSAE on the decoder features of SAEs, following the methodology of \citet{leask2025sparse}. The MetaSAE follows the same training procedure as ordinary SAEs (Sec.~\ref{subsec:Experimental Setup}) but operates on decoder features rather than LLM activations, attempting to decompose these higher-level representations into more atomic latent components (detailed descriptions provided in Appx. \ref{app:MetaSAE-Based Feature Composition metrics details}). The MetaSAE employs a BatchTopK architecture with sparsity $k = 4$ and a dictionary size reduced to 25\% of the original SAEs. We measure composition via explained variance, where lower values indicate higher atomicity. Fig.~\ref{fig:atomicity_metrics}a shows that OrtSAE attains a composition rate much lower (by $0.06$ at $\text{L}0$ of $70$) than BatchTopK SAE and similar to Matryoshka SAE, reflecting pronounced feature atomicity and resistance to decomposition into simpler constituents.

\paragraph{Feature Absorption Analysis.} Feature absorption is evaluated using SAEBench~\citep{karvonen2025saebench}. Following established methodologies for studying absorption in sparse autoencoders~\citep{chanin2024absorption}, we perform tests in the first-letter classification and hierarchical concept domains (detailed descriptions provided in Appx. B). Fig.~\ref{fig:atomicity_metrics}b shows that OrtSAE achieves a significantly reduced absorption rate compared to BatchTopK SAE (by $0.17$ at $\text{L}0$ of $70$), but slightly higher than Matryoshka SAE, indicating effective minimization of conceptual overlap compared to traditional SAEs.

\paragraph{Clustering Properties of Decoder Features.} The clustering coefficient measures how tightly connected features are in a graph of decoder interactions, where edges represent high cosine similarity between features (detailed descriptions provided in Appx. C). A lower coefficient indicates that features are more independent, forming fewer interconnected groups, which suggests greater feature atomicity. We evaluate this across 10 similarity thresholds at varying edge densities (the ratio of existing edges to the number of all possible edges). Fig.~\ref{fig:atomicity_metrics}c shows the clustering coefficient of OrtSAE is substantially lower than that of BatchTopK SAE and similar to Matryoshka SAE, signifying enhanced feature independence compared to traditional SAEs.

\paragraph{Cross-Model Feature Overlap Analysis.} We measure feature uniqueness by computing maximum pairwise cosine similarity between OrtSAE and BatchTopK SAE features, at a $\text{L}0$ of $70$. A feature is considered unique if all cross-model similarities are below 0.2. OrtSAE retains 9\% unique features, compared to 1.5\% for BatchTopK (Fig.~\ref{fig:atomicity_metrics}d). This six-fold increase in unique features highlights OrtSAE’s ability to discover novel features, enhancing scalability for larger dictionaries.

\paragraph{Qualitative Analysis of Feature Atomicity.} We analyze feature atomicity by decomposing BatchTopK SAE features into sparse combinations of OrtSAE features, following a methodology adapted from MetaSAE \citep{leask2025sparse}. We select BatchTopK SAE (at $\text{L}0$ of $70$) features if their OrtSAE (at $\text{L}0$ of $70$) approximation has a cosine similarity above 0.95 and each coefficient is at least 0.1, ensuring meaningful contributions. Fig.~\ref{fig:queen_example} illustrates the decomposition of a BatchTopK SAE feature for ``Queen terms'' into two OrtSAE features: one for ``Queen terms'' and another for ``titles and royal concepts''. This demonstrates how OrtSAE disentangles broader concepts absorbed by specialized features. Additional examples of feature decompositions are provided in Appx. D.

These results demonstrate OrtSAE improves feature atomicity over traditional SAEs, achieving lower composition and absorption rates, reduced clustering, and a higher proportion of unique features. Qualitative evidence supports these findings, demonstrating the effective decomposition of complex BatchTopK features by OrtSAE features. These results highlight the efficacy of orthogonality constraint in yielding disentangled representations, paving the way for downstream tasks.

\subsection{Downstream benchmarks}
\label{subsec:Downstream benchmarks}
We evaluate OrtSAE using the SAEBench \citep{karvonen2025saebench}, which measures SAEs quality across a diverse set of tasks related to practical downstream applications. The key metrics we report on are Spurious Correlation Removal (SCR), Targeted Probe Perturbation (TPP), Sparse Probing, and RAVEL.

\begin{figure}[t!]
    \centering
    \begin{subfigure}[t]{0.243\textwidth}
        \centering
        \includegraphics[width=\textwidth]{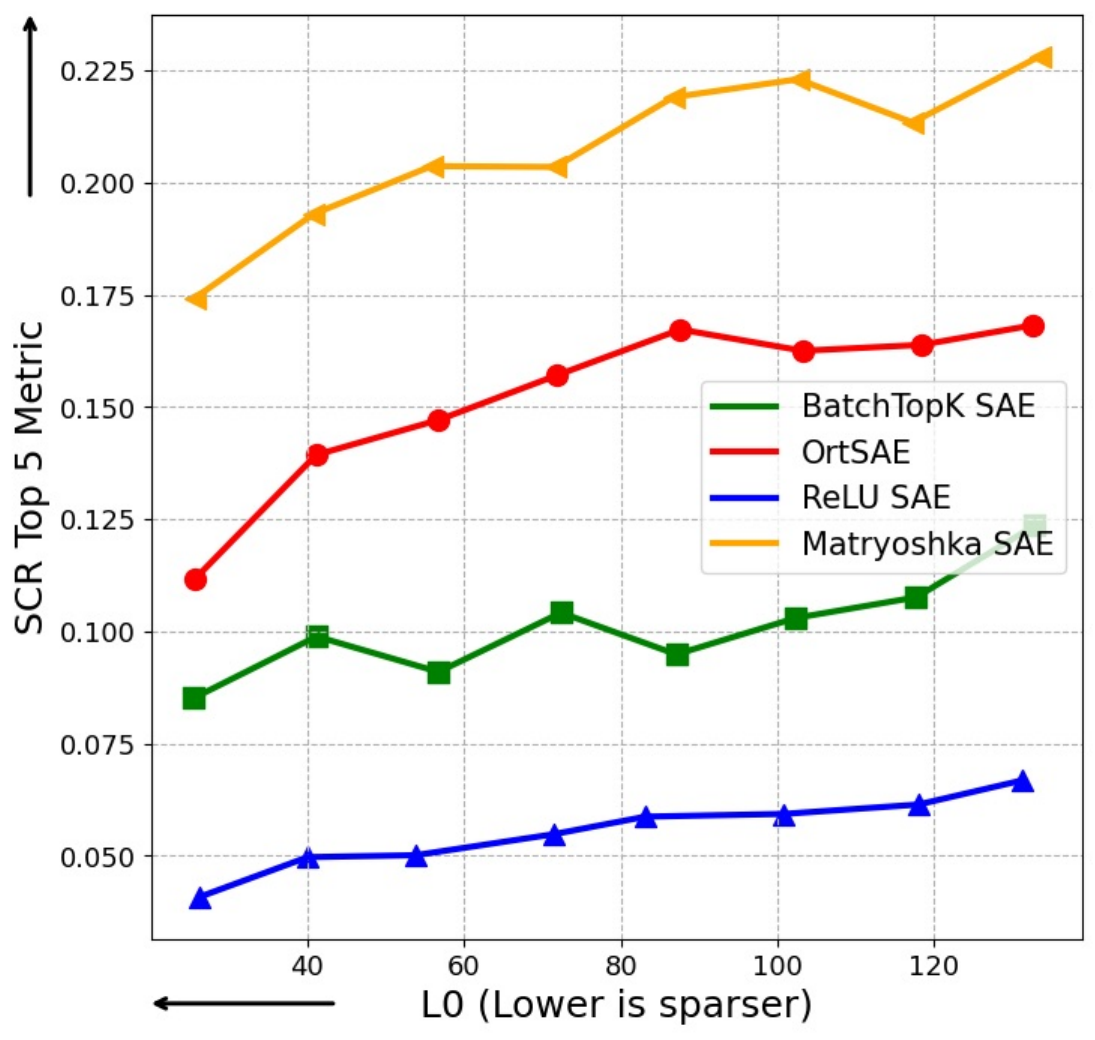}
        \caption{SCR score.}
        \label{fig:ce_loss_score}
    \end{subfigure}
    \begin{subfigure}[t]{0.243\textwidth}
        \centering
        \includegraphics[width=\textwidth]{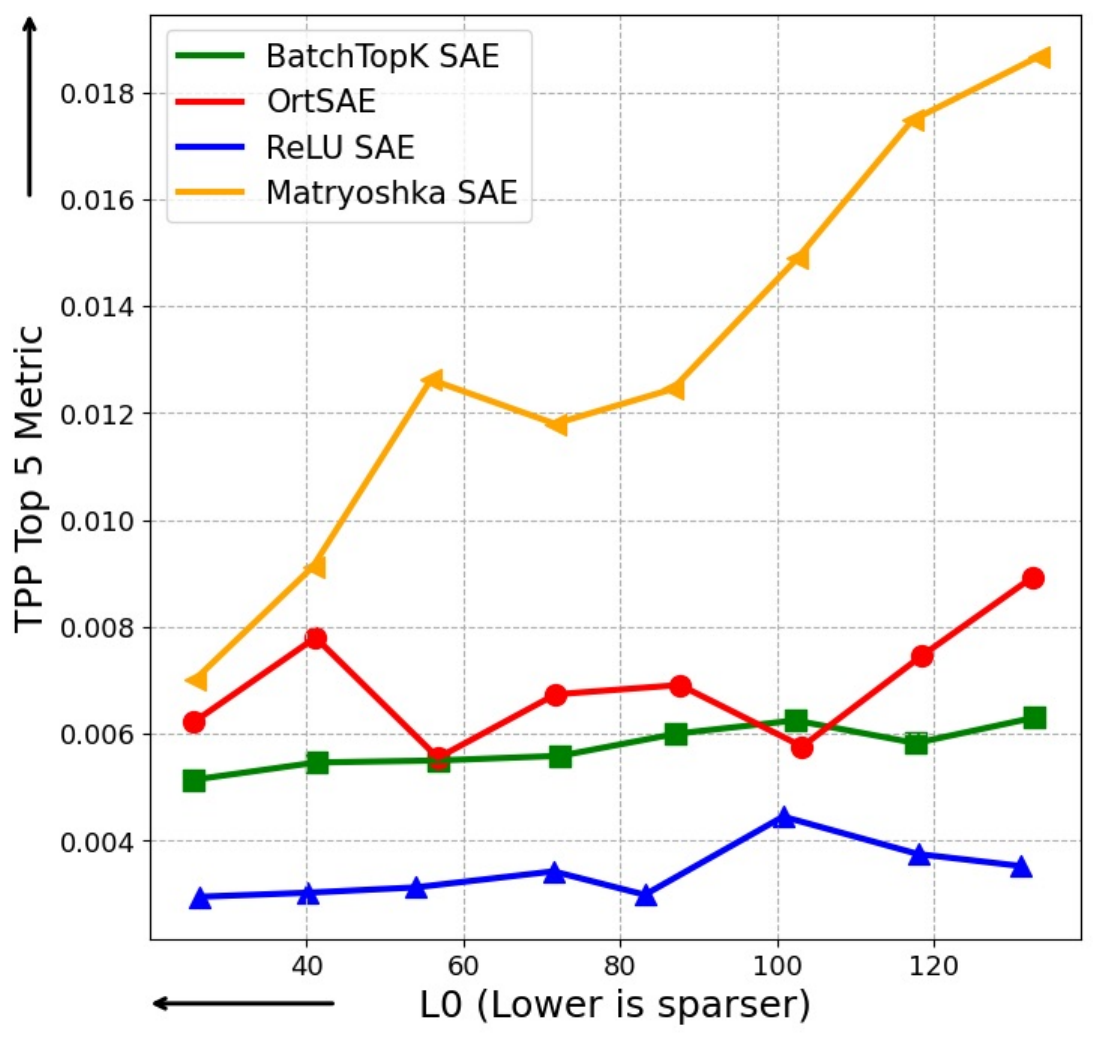}
        \caption{TPP score.}
        \label{fig:kl_div_score}
    \end{subfigure}
    \begin{subfigure}[t]{0.243\textwidth}
        \centering
        \includegraphics[width=\textwidth]{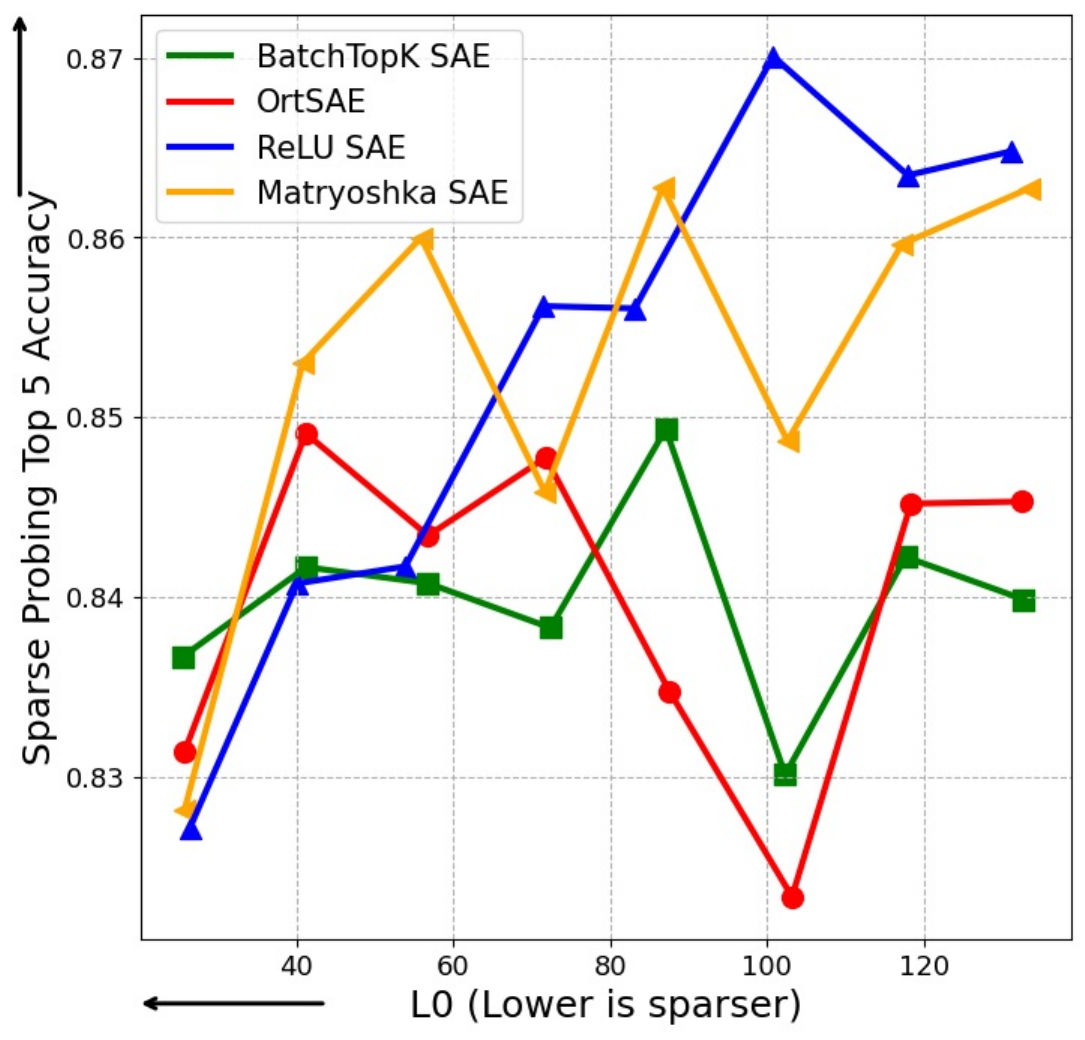}
        \caption{Probing accuracy.}
        \label{fig:kl_div_score}
    \end{subfigure}
    \begin{subfigure}[t]{0.243\textwidth}
        \centering
        \includegraphics[width=\textwidth]{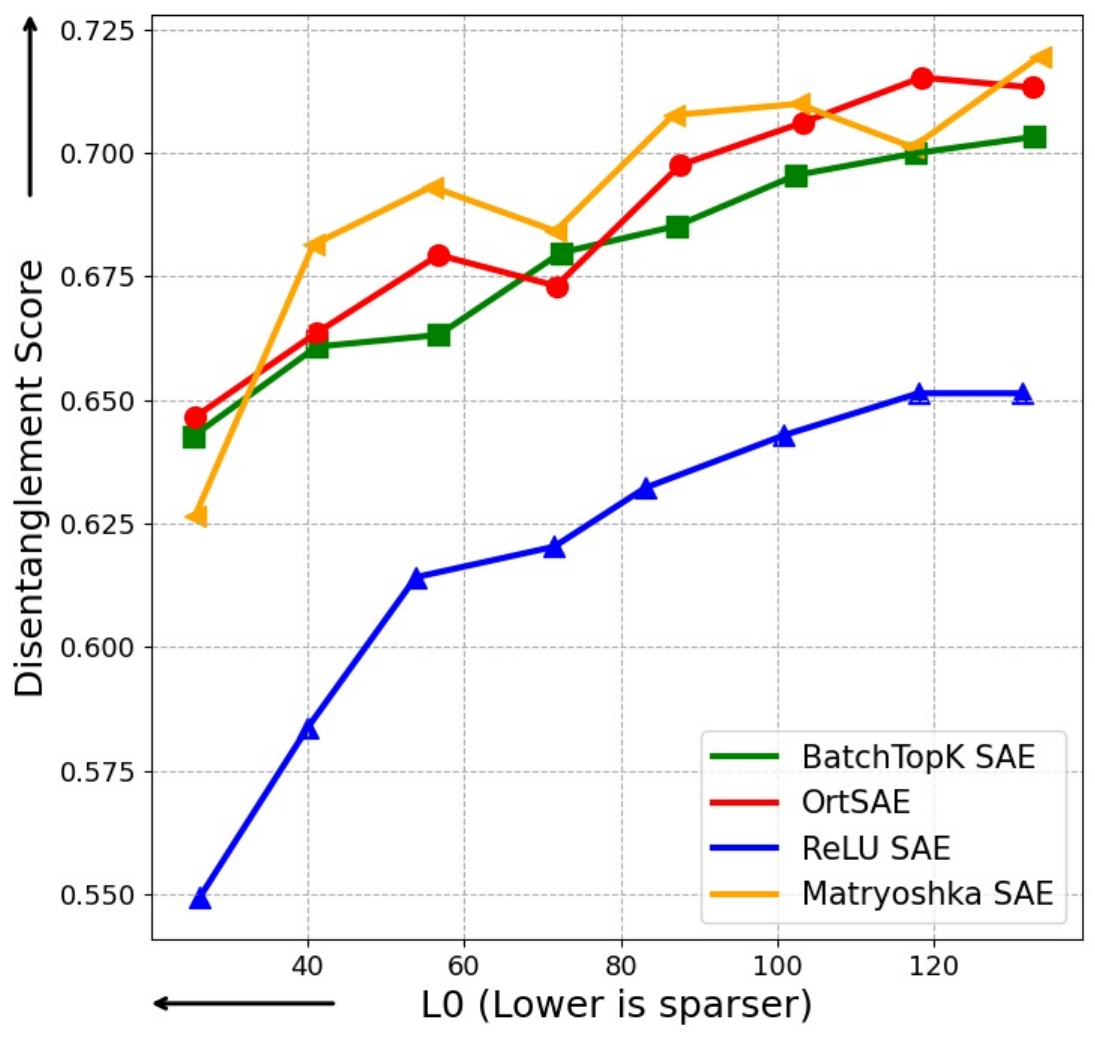}
        \caption{RAVEL score.}
        \label{fig:}
    \end{subfigure}
    \caption{\textbf{Results on SAEBench.} (a) Spurious Correlation Removal: OrtSAE outperforms traditional SAEs but achieves lower scores than Matryoshka. (b) Targeted Probe Perturbation: OrtSAE shows modest improvements over traditional SAEs, while Matryoshka demonstrates the strongest performance. (c) Sparse Probing: OrtSAE maintains accuracy comparable to baseline methods. (d) RAVEL: OrtSAE achieves scores similar to both BatchTopK and Matryoshka SAEs.}
    \label{fig:scr_tpp}
\end{figure}

\paragraph{Spurious Correlation Removal and Targeted Probe Perturbation.} SCR evaluates the removal of spurious correlations (e.g., gender in profession classification) by zero-ablating SAE latents, while TPP tests class-specific concept isolation in multi-class settings via targeted ablation. Fig.~\ref{fig:scr_tpp} demonstrates that OrtSAE achieves stronger performance than traditional SAEs in both tasks, with a significant improvement in SCR scores and modest gains in TPP compared to BatchTopK SAE. While Matryoshka SAE achieves the highest absolute scores, OrtSAE delivers similarly strong results with lower computational overhead and traditional architecture.

\paragraph{Sparse Probing.} Sparse Probing evaluates the ability of SAEs to isolate specific concepts, such as sentiment, within individual latents without explicit supervision. For each concept, we select the top-\textit{k} latents by comparing their mean activations on positive versus negative examples. A linear probe is then trained on these latents to predict the concept. High probe accuracy indicates that the latents effectively capture the target concept in a disentangled manner. As shown in Fig.~\ref{fig:scr_tpp}c, OrtSAE achieves probing accuracy comparable to BatchTopK and Matryoshka SAEs across various sparsity levels, demonstrating its capability to produce interpretable, concept-aligned features.

% \paragraph{Sparse Probing.} Sparse Probing measures how well SAEs align individual latents with defined concepts (e.g., sentiment) using linear probes on the top-\textit{k} latents. Fig.~\ref{fig:scr_tpp}c shows OrtSAE, BatchTopK SAE, and Matryoshka SAE demonstrate similar performance, indicating consistent concept alignment across all models.

\paragraph{RAVEL.} RAVEL (Resolving Attribute–Value Entanglements in Language
Models) \citep{huang2024ravel} evaluates disentanglement by manipulating attributes in LLM activations (e.g., transferring city-related features from “Tokyo” to “Paris”) while minimizing interference with unrelated attributes (e.g., France-related context). Fig.~\ref{fig:scr_tpp}d demonstrates that OrtSAE, BatchTopK SAE, and Matryoshka SAE exhibit equivalent performance, highlighting OrtSAE’s ability to enable precise interventions without compromising efficacy.

Evaluated against traditional SAEs, OrtSAE demonstrates superior SCR performance, competitive TPP results, and consistently comparable outcomes in Sparse Probing and RAVEL, highlighting its ability to generate atomic, disentangled features without sacrificing downstream utility.

% \section{Limitations}
% \label{sec:limitations}

% Our evaluation was confined to layer 12 of the Gemma-2-2B model, which may constrain the generalizability of our findings to other layers or different model architectures. While the superposition hypothesis suggests potential applicability across models, further experimentation is necessary to verify the sustainability of our approach. Our reliance on quantitative and automated metrics, though rigorous, may not fully capture the nuances of human-centric interpretability, potentially overlooking more complex feature interpretation challenges. Additionally, we observe that our new training procedure resulted in a higher proportion of dead features (approximately 6\% compared to 1\% in standard architectures), suggesting possible under-utilization of latent capacity. Furthermore, OrtSAE introduces additional hyperparameters, such as the orthogonality coefficient $\gamma$ and the number of chunks $K(m)$, which require careful tuning to balance orthogonality enforcement and reconstruction performance. This additional complexity in the training process represents a practical challenge for implementation. These limitations highlight areas for future research to enchance the robustness and broaden the applicability of OrtSAE. 

\section{Conclusion}
\label{sec:Discussion and Conclusion}

In this work, we introduce OrtSAE, a novel sparse autoencoder training approach that enhances latent atomicity through orthogonal constraints on decoder features. This method effectively addresses feature absorption and composition, key obstacles to interpretable representations, while preserving reconstruction fidelity comparable to traditional SAEs. Notably, OrtSAE achieves this with minimal computational overhead through an efficient chunk-wise orthogonalization penalty that scales linearly with feature count. Experiments across different language model families demonstrate OrtSAE's significant reductions in absorption and composition rates, yielding 9\% more distinct features, superior spurious correlation removal (+6\%), and on-par performance across other SAEBench tasks. These insights underscore geometric constraints' role in disentangling superposed representations, offering fresh perspectives on the superposition hypothesis. Future work should investigate using orthogonal features as more interpretable building blocks for neural circuit discovery, potentially leading to clearer mechanistic models of model computations.

\section{Reproducibility statement}

To ensure the reproducibility of our work, we have taken several measures throughout the paper and supplementary materials. All experimental details, including model architectures, hyperparameters, and training procedures, are comprehensively documented in Section \ref{subsec:Experimental Setup} and Appendix \ref{app:sae_details}. We provide complete specifications for our OrtSAE implementation, including the orthogonal penalty formulation and chunking strategy in Section \ref{subsec:ortsae_training}. The evaluation metrics and benchmarks are described in detail in Sections \ref{subsec:Fundamental Performance Metrics}-\ref{subsec:Downstream benchmarks}, with additional methodological explanations in Appendices \ref{app:additional_experiments}-\ref{app:MetaSAE-Based Feature Composition metrics details}. Code for OrtSAE training, evaluation scripts, feature analysis tools, and SAEBench integration will be released anonymously as supplementary material and made fully public upon acceptance. The datasets used in our experiments (OpenWebText) are publicly available, and we specify exact data processing steps in Appendix {\ref{app:sae_details}}. For the SAEBench evaluations, we follow established protocols from prior work with detailed descriptions of each task. Additional experiments on different model architectures and layers (Appendix \ref{app:additional_experiments}) and ablation studies on key hyperparameters (Appendix \ref{app:chunk_analysis}) further validate the robustness of our approach.

\bibliographystyle{plainnat}
% \bibliography{references}

\appendix

\section{Additional details of SAE training setup}
\label{app:sae_details}
Following the approach of \citep{bussmann2025learning}, we train Sparse Autoencoders (SAEs) on activations from layer 12 of the Gemma-2-2B model \citep{team2024gemma} (26 layers total) to align with prior work. To assess the generalizability of our approach, we also conduct experiments on layer 20 of Gemma-2-2B and layer 20 of Llama-3-8B \citep{dubey2024llama} (32 layers total). Each SAE has latent space of size $m = 65536$ and sparsity levels $\text{L}0$ in $\{25, 40, 55, 70, 85, 100, 115, 130\}$. The training uses 500 million tokens from the OpenWebText dataset \citep{gokaslan2019} with a context length of 1024. 
We employ the AdamW optimizer \citep{loshchilov2017decoupled} with a learning rate of $2 \times 10^{-4}$ and a batch size of $2048$.
All SAE variants except ReLU SAE employ an auxiliary loss coefficient $\alpha = 1/32$ to mitigate dead features during training.
As a basis of OrtSAE, we follow BatchTopK SAE \href{https://github.com/saprmarks/dictionary_learning.git}{repository}, leveraging BatchTopK’s precise $\text{L}0$ sparsity control.
For OrtSAE, we set the number of chunks $K(m) = \lceil m / 8192 \rceil$ (yielding $K = 8$ for $m = 65536$) with $\gamma = 0.25$.
All SAEs are trained comparably with identical data ordering and hyperparameters on $1$ $\text{H}100$ GPU, $80\text{GB}$. For one SAE, training requires approximately $10$ hours. To ensure reproducibility, we will publicly release all code, hyperparameters, and instructions for accessing the datasets. The results of additional experiments on layer 20 of Gemma-2-2B and Llama-3-8B focusing on a sparsity level of $\text{L}0=70$ are reported in the Appendix \ref{app:additional_experiments}.

\section{Additional Experiments on Gemma-2-2B and Llama-3-8B}
\label{app:additional_experiments}

To address the generalizability of our findings, we conducted additional experiments on layer 20 of Gemma-2-2B (26 layers total) and layer 20 of Llama-3-8B (32 layers total) \citep{dubey2024llama}, following the experimental setup described in Sec.~\ref{subsec:Experimental Setup}. We trained SAEs with a sparsity level of $\text{L}0=70$ and measured key metrics: explained variance, mean cosine similarity, composition rate, absorption rate, and Spurious Correlation Removal (SCR) score. The results, presented in Tables~\ref{tab:gemma_layer20} and \ref{tab:llama_layer20}, confirm the findings from layer 12 of Gemma-2-2B, showing consistent reductions in feature absorption and composition, as well as improved performance in tasks such as Spurious Correlation Removal. Notably, the Explained Variance gap between Matryoshka SAE and OrtSAE widens to 0.04 in Llama-3-8B (0.722 vs. 0.762), reinforcing OrtSAE's advantage in reconstruction performance.

\begin{table}[h]
    \centering
    \caption{\textbf{Performance of SAEs trained on layer 20 of Gemma-2-2B with L0=70.}}
    \vspace{5pt}
    \small
    \begin{tabular}{@{\hskip 0.3cm}l|c|c|c|c|c@{\hskip 0.3cm}}
        \toprule
        \textbf{SAE model} & \textbf{Expl. var.} & \textbf{Mean Cos. sim.} & \textbf{Comp. rate} & \textbf{Abs. rate} & \textbf{SCR score} \\
        \midrule
        ReLU SAE & 0.784 & 0.549 & 0.527 & 0.371 & 0.144 \\
        BatchTopK SAE & \textbf{0.843} & 0.354 & 0.490 & 0.220 & 0.308 \\
        Matryoshka SAE & 0.811 & 0.148 & 0.349 & \textbf{0.015} & \textbf{0.385} \\
        OrtSAE & 0.836 & \textbf{0.112} & \textbf{0.340} & 0.095 & 0.322 \\
        \bottomrule
    \end{tabular}
    \label{tab:gemma_layer20}
\end{table}

\begin{table}[h]
    \centering
    \caption{\textbf{Performance of SAEs trained on layer 20 of Llama-3-8B with L0=70.}}
    \vspace{5pt}
    \small
    \begin{tabular}{@{\hskip 0.3cm}l|c|c|c|c|c@{\hskip 0.3cm}}
        \toprule
        \textbf{SAE model} & \textbf{Expl. var.} & \textbf{Mean Cos. sim.} & \textbf{Comp. rate} & \textbf{Abs. rate} & \textbf{SCR score} \\
        \midrule
        ReLU SAE & 0.704 & 0.517 & 0.413 & 0.490 & 0.060 \\
        BatchTopK SAE & \textbf{0.769} & 0.327 & 0.461 & 0.148 & 0.103 \\
        Matryoshka SAE & 0.722 & 0.149 & 0.323 & \textbf{0.022} & \textbf{0.191} \\
        OrtSAE & 0.762 & \textbf{0.107} & \textbf{0.316} & 0.070 & 0.151 \\
        \bottomrule
    \end{tabular}
    \label{tab:llama_layer20}
\end{table}

% Training times, summarized in Table~\ref{tab:training_times}, show that the modified OrtSAE introduces small overhead (e.g., 1.11$\times$ the time of BatchTopK at $K=8$), but reaches near-parity at $K=64$ (1.04$\times$). Overall, OrtSAE demonstrates both robustness and scalability, maintaining core SAE performance across varying chunk counts and achieving near-identical effectiveness with the modified approach.

\section{Effect of Number of Chunks on OrtSAE Performance}
\label{app:chunk_analysis}

We evaluated the impact of the number of chunks and the frequency of orthogonality loss computation on OrtSAE performance, following the experimental setup in Sec.~\ref{subsec:Experimental Setup}. OrtSAE was tested with chunk counts $K \in \{4, 8, 16, 32, 64\}$, using a fixed sparsity at $\mathrm{L}0$ of $70$. Additionally, to enhance computational efficiency, we explored a modified OrtSAE variant where the orthogonality loss is computed every fifth training iteration, with the orthogonality coefficient $\gamma$ scaled by a factor of 5 to maintain regularization strength.

\begin{figure}[h]
    \centering
    \begin{subfigure}[t]{0.243\textwidth}
        \centering
        \includegraphics[width=\textwidth]{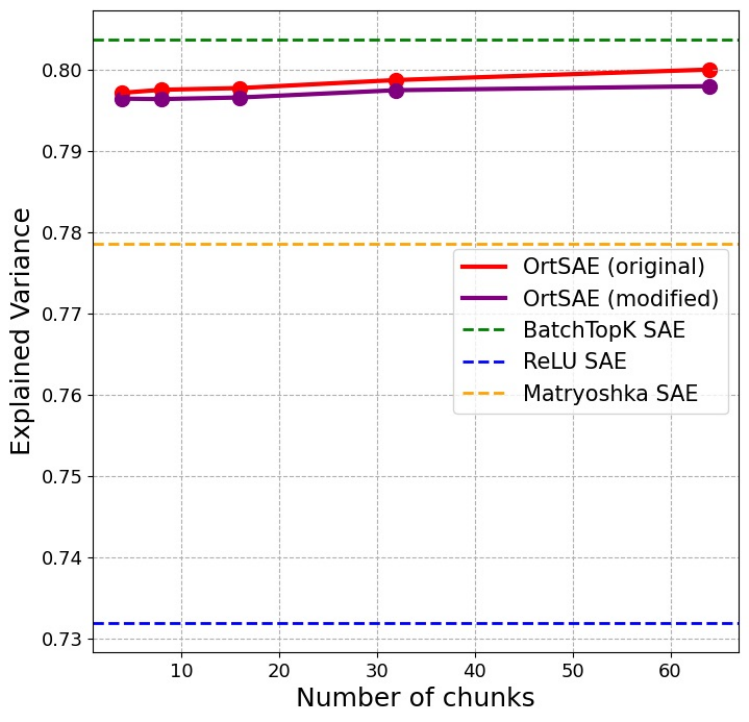}
        \caption{Explained variance}
        \label{fig:ce_loss_score}
    \end{subfigure}
    \begin{subfigure}[t]{0.243\textwidth}
        \centering
        \includegraphics[width=\textwidth]{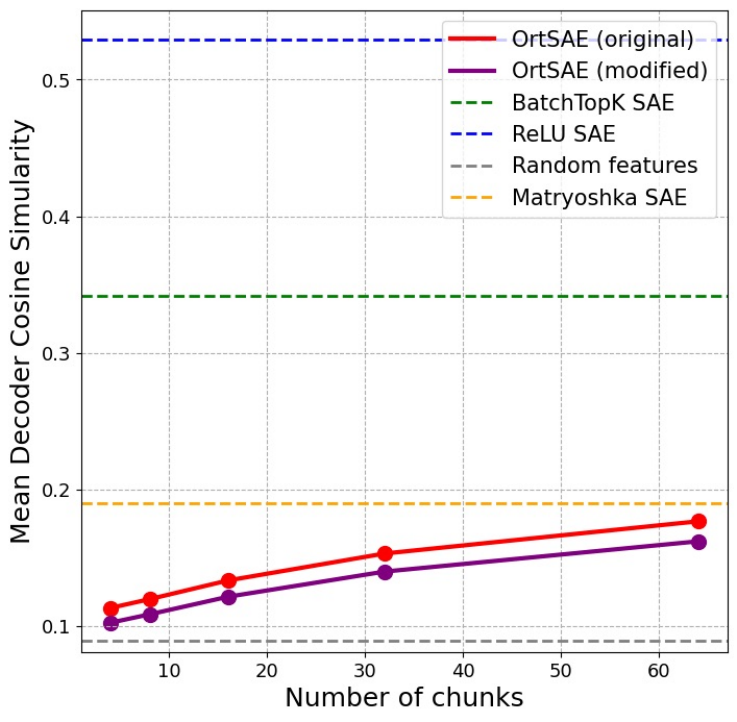}
        \caption{Mean cosine similarity}
        \label{fig:kl_div_score}
    \end{subfigure}
    \begin{subfigure}[t]{0.243\textwidth}
        \centering
        \includegraphics[width=\textwidth]{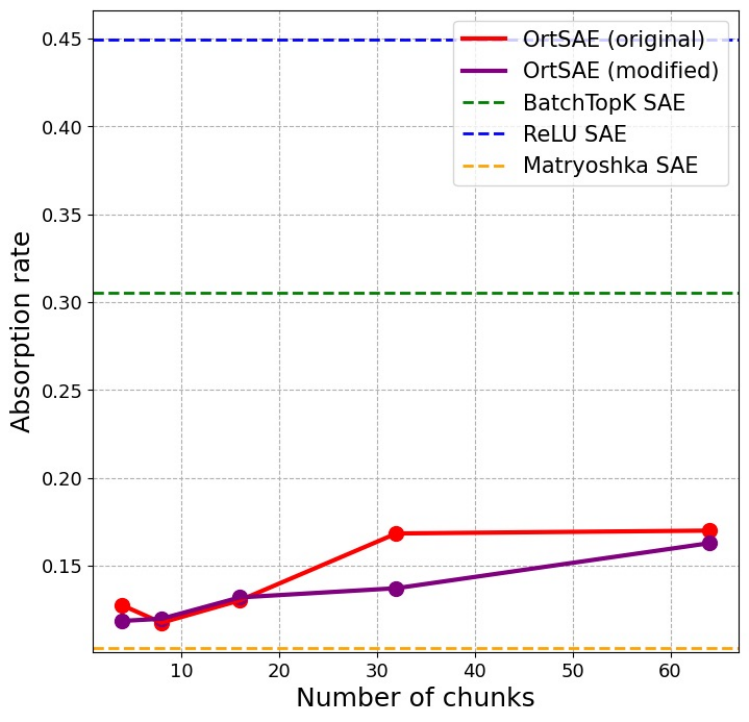}
        \caption{Absorption rate}
        \label{fig:kl_div_score}
    \end{subfigure}
    \begin{subfigure}[t]{0.243\textwidth}
        \centering
        \includegraphics[width=\textwidth]{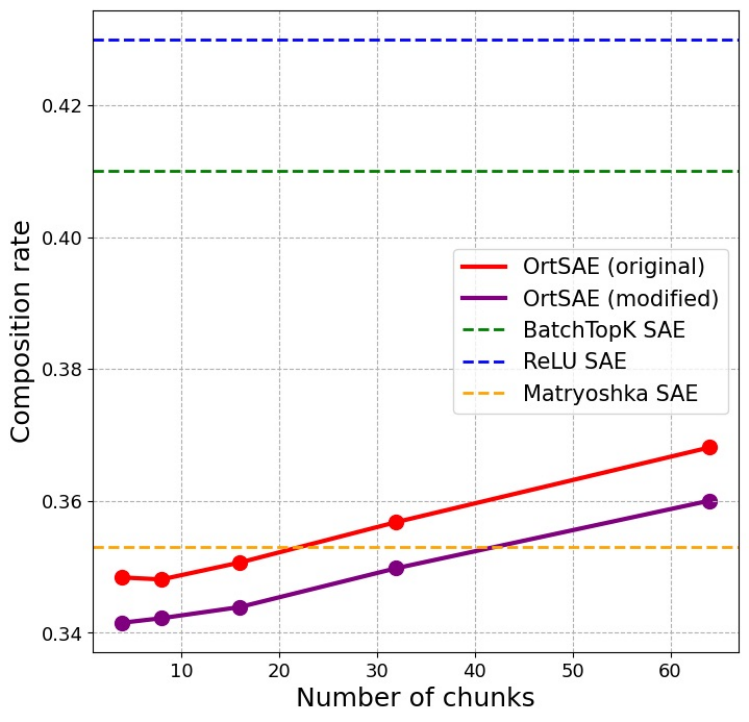}
        \caption{Composition rate}
        \label{fig:}
    \end{subfigure}
    \caption{\textbf{Performance of OrtSAE across different number of chunks at L0 of 70}. OrtSAE shows robust performance in core reconstruction and atomicity metrics across different number of chunks.}
    \label{fig:add_experiments}
\end{figure}

As shown in Figure~\ref{fig:add_experiments}, OrtSAE demonstrates robust performance in core reconstruction and atomicity metrics across different numbers of chunks. The original OrtSAE and the modified OrtSAE show similar trends, with the modified version maintaining performance even better then original. Explained variance (Fig.\ref{fig:add_experiments}a) remains stable around 0.77–0.78 across chunk counts for both variants, slightly outperforming ReLU SAE and Matryoshka SAE. Mean cosine similarity (Fig.\ref{fig:add_experiments}b) increases modestly with more chunks (e.g., from 0.10 at $K=4$ to 0.20 at $K=64$), but remains lower than BatchTopK and ReLU SAEs. Absorption rate (Fig.\ref{fig:add_experiments}c) and composition rate (Fig.\ref{fig:add_experiments}d) also increase slightly with more chunks (e.g., absorption from 0.15 to 0.20), yet both variants outperform traditional SAEs. The modified OrtSAE achieves performance within 1–2\% of the original OrtSAE across these metrics, reducing the computational overhead of the orthogonality loss by approximately five times, as it is calculated in only 20\% of training iterations. Overall, OrtSAE demonstrates both robustness and scalability by maintaining core SAE performance across varying chunk counts while reducing training overhead to within $4\%$ of the BatchTopK baseline (see Table \ref{tab:training_times}).

\begin{table}[h]
    \centering
    \caption{\textbf{Training times for SAE variants.} Ratios are relative to BatchTopK SAE.}
    \vspace{10pt}
    \begin{tabular}{l|c|c}
        \toprule
        \textbf{Method} & \textbf{Training Time (minutes)} & \textbf{Time Ratio*} \\
        \midrule
        BatchTopK SAE & 325 & 1.0$\times$ \\
        Matryoshka SAE & 373 & 1.15$\times$ \\
        OrtSAE modified ($K=8$) & 361 & 1.11$\times$ \\
        OrtSAE modified ($K=64$) & \textbf{340} & \textbf{1.04$\times$} \\
        \bottomrule
        \multicolumn{3}{l}{\footnotesize *Relative to BatchTopK SAE}
    \end{tabular}
    \label{tab:training_times}
\end{table}

\section{KL-divergence score definition}
\label{app:KL-divergence score definition}

The KL-divergence score assesses how effectively a sparse autoencoder (SAE) preserves the predictive behavior of a language model by comparing next-token probability distributions. We define $P_{\text{orig}}$ as the distribution from the original model, $P_{\text{SAE}}$ as the distribution when activations are replaced by SAE reconstructions, and $P_{\text{ablated}}$ as the distribution when activations are set to zero, serving as a baseline.

The score is given by:
\[
\text{KL-Divergence Score} = \frac{D_{KL}(P_{\text{ablated}} \parallel P_{\text{orig}}) - D_{KL}(P_{\text{SAE}} \parallel P_{\text{orig}})}{D_{KL}(P_{\text{ablated}} \parallel P_{\text{orig}})}
\]
where $D_{KL}$ denotes the Kullback-Leibler divergence. This metric ranges from 0, indicating no improvement over the zero-ablated baseline, to 1, indicating perfect reconstruction ($P_{\text{SAE}} = P_{\text{orig}}$). In SAEBench, the score is averaged over a dataset to evaluate SAE reconstruction quality.

\section{LogLoss Results}
\label{app:logloss}

To further evaluate the impact of decoded activations on the base language model’s predictive performance, we compute LogLoss scores for SAEs trained on layer 12 of Gemma-2-2B across various sparsity levels ($\text{L}0 \in \{40, 70, 100, 130\}$). LogLoss measures the negative log-likelihood of the model’s next-token predictions, with lower values indicating better preservation of the base model’s predictive behavior. The results, shown in Table~\ref{tab:logloss}, align closely with the KL-divergence findings (Appendix~\ref{app:KL-divergence score definition}), confirming that OrtSAE maintains predictive performance comparable to BatchTopK SAE, with both outperforming ReLU SAE and Matryoshka SAE.

\begin{table}[h]
    \centering
    \caption{\textbf{LogLoss of SAEs trained on layer 12 of Gemma-2-2B.} Lower LogLoss indicates better preservation of the base language model’s predictive performance.}
    \vspace{5pt}
    \small
    \begin{tabular}{@{\hskip 0.3cm}l|c|c|c|c@{\hskip 0.3cm}}
        \toprule
        \textbf{SAE model} & \textbf{L0=40} & \textbf{L0=70} & \textbf{L0=100} & \textbf{L0=130} \\
        \midrule
        No SAE (LLM’s LogLoss) & 2.4533 & 2.4533 & 2.4533 & 2.4533 \\
        ReLU SAE & 2.7657 & 2.6562 & 2.6094 & 2.5935 \\
        BatchTopK SAE & \textbf{2.5623} & \textbf{2.5312} & \textbf{2.5152} & \textbf{2.5020} \\
        Matryoshka SAE & 2.5786 & 2.5321 & 2.5161 & 2.5025 \\
        OrtSAE & 2.5646 & 2.5319 & 2.5159 & 2.5024 \\
        \bottomrule
    \end{tabular}
    \label{tab:logloss}
\end{table}

\section{Feature Interpretability Metrics Details}
\label{app:autointerp}

To evaluate the interpretability of Sparse Autoencoder (SAE) features, we follow the automated interpretability methodology outlined in \citep{paulo2024automatically, karvonen2025saebench}, leveraging large language models (LLMs) to generate and validate human-readable feature descriptions. We select 1,000 random SAE features, excluding "dead" features. For each feature, we collect up to 10 top-activating sequences from the OpenWebText dataset \citep{gokaslan2019} and prompt GPT-4o-mini to generate a concise description, such as "sentiment terms" or "math expressions," capturing the feature’s core concept.

To assess these descriptions, we create a test set for each feature with 100 sequences: 50 activating the feature at varying strengths and 50 random non-activating sequences, all sourced from OpenWebText. A separate GPT-4o-mini model predicts whether each sequence activates the feature based on the description, treating it as a binary (yes/no) classification task. The \textit{Autointerp Score}, shown in Fig.~\ref{fig:core_metrics}d, is the prediction accuracy, measuring how well the description generalizes to new data. A high score indicates a monosemantic, interpretable feature, while a lower score may suggest polysemanticity or an inaccurate description. OrtSAE demonstrates interpretability comparable to BatchTopK and Matryoshka SAEs across sparsity levels (Sec.~\ref{subsec:Fundamental Performance Metrics}), confirming its ability to produce clear, disentangled feature representations.

\section{MetaSAE-Based Feature Composition Metrics Details}
\label{app:MetaSAE-Based Feature Composition metrics details}

The MetaSAE-based feature composition analysis, originally proposed by \citet{leask2025sparse} and extended by \citet{bussmann2025learning}, provides a quantitative method for assessing the atomicity of features learned by sparse autoencoders. This approach measures the degree of feature composition by training a secondary sparse autoencoder (MetaSAE) on the feature vectors of the primary SAE to decompose them into more atomic components.

In our implementation, we train MetaSAEs on the decoder weight matrices of both OrtSAE and BatchTopK SAE. The decoder weights are treated as input data points, where each feature vector from the primary SAE's dictionary serves as an input to the MetaSAE. The MetaSAE follows the same BatchTopK architecture as our primary SAEs but operates on this different input space.

The MetaSAE is configured with a dictionary size equal to one-quarter of the primary SAE's dictionary size (16,384 meta-latents for our primary SAEs with 65,536 features). We apply a sparsity constraint ensuring an average of 4 active meta-latents per decoder vector reconstruction. The MetaSAE learns to reconstruct each primary SAE feature vector using a sparse combination of meta-latents, where the meta-features represent atomic sub-components of the primary SAE's features.

The key metric for assessing composition is the explained variance, which measures the proportion of variance in the primary SAE's decoder weights that the MetaSAE can reconstruct. A higher explained variance indicates that the MetaSAE can effectively reconstruct the primary SAE's features using shared, atomic sub-features, suggesting that the original features were composed of these simpler components. Conversely, a lower explained variance implies that the primary features are already atomic and resist decomposition into simpler constituents. In our experiments, we interpret lower MetaSAE explained variance as indicating better feature atomicity in the primary SAE.

This methodology provides an objective, quantitative measure of feature composition that complements our qualitative analyses and other atomicity metrics, offering insights into the hierarchical structure of the representations learned by different SAE variants.

\section{Clustering Coefficient Definition}
\label{app:clustering_coefficient}

% Defining the global clustering coefficient
The global clustering coefficient quantifies the tendency of nodes in a graph to form clusters. For Sparse Autoencoders (SAEs), nodes represent decoder features, and edges connect feature pairs with cosine similarity above a threshold.

% Specifying the edge density
Edge density, the proportion of possible edges present, is defined as:

\[
\text{density} = \frac{2E}{n(n-1)},
\]

where \(E\) is the number of edges and \(n\) is the number of nodes. Varying the similarity threshold adjusts the density, enabling analysis across connectivity levels.

% Providing the mathematical definition
The clustering coefficient \(C\) is computed as:

\[
C = \frac{3 \times \text{number of triangles}}{\text{number of connected triples}},
\]

where a triangle is three nodes fully connected by edges, and a connected triple is three nodes linked by at least two edges. \(C\) ranges from 0 (no clustering) to 1 (maximal clustering).

\section{Additional Qualitative Examples}
\label{app:qualitative_examples}

We provide three additional examples of BatchTopK SAE features decomposed into orthogonal, atomic OrtSAE components. These cases further illustrate how OrtSAE disentangles composite concepts through orthogonality constraint.

\begin{figure}[h!]
    \centering
    \begin{subfigure}[t]{0.32\textwidth}
        \centering
        \includegraphics[width=\textwidth]{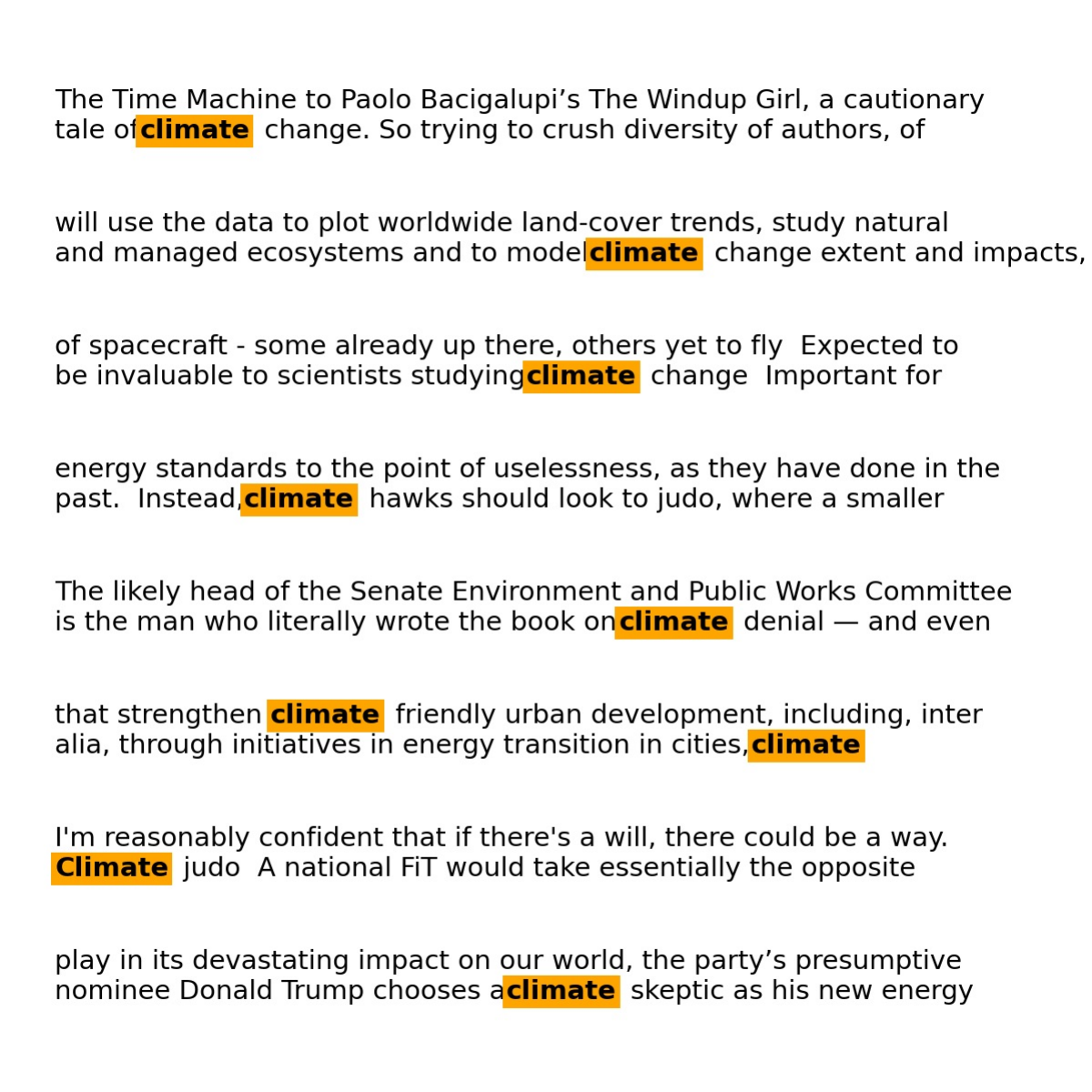}
        \caption{BatchTopK SAE feature activating on climate-related terms.}
        \label{fig:climate_0}
    \end{subfigure}
    \hfill
    \begin{subfigure}[t]{0.32\textwidth}
        \centering
        \includegraphics[width=\textwidth]{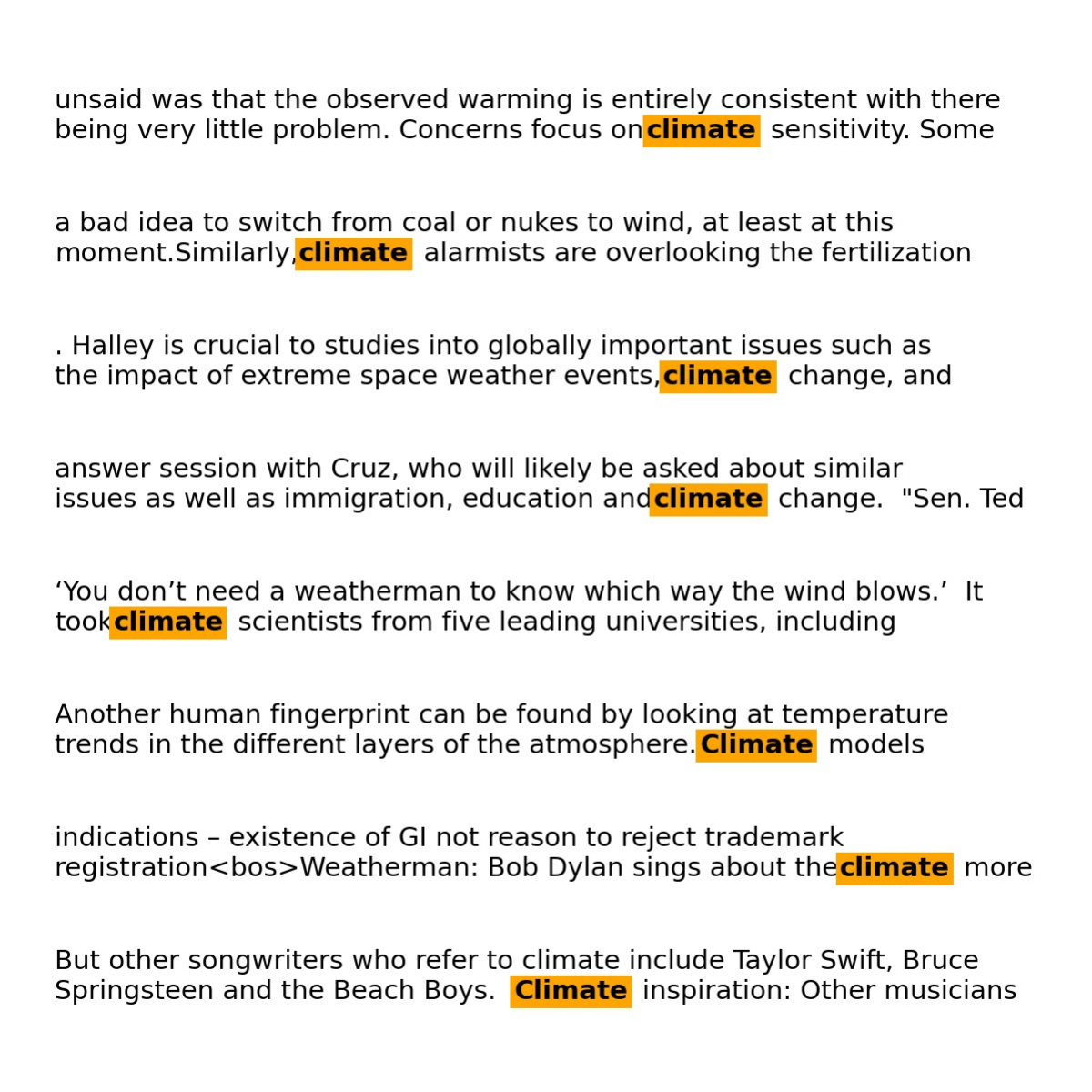}
        \caption{OrtSAE feature activating on on climate-related terms.}
        \label{fig:climate_1}
    \end{subfigure}
    \hfill
    \begin{subfigure}[t]{0.32\textwidth}
        \centering
        \includegraphics[width=\textwidth]{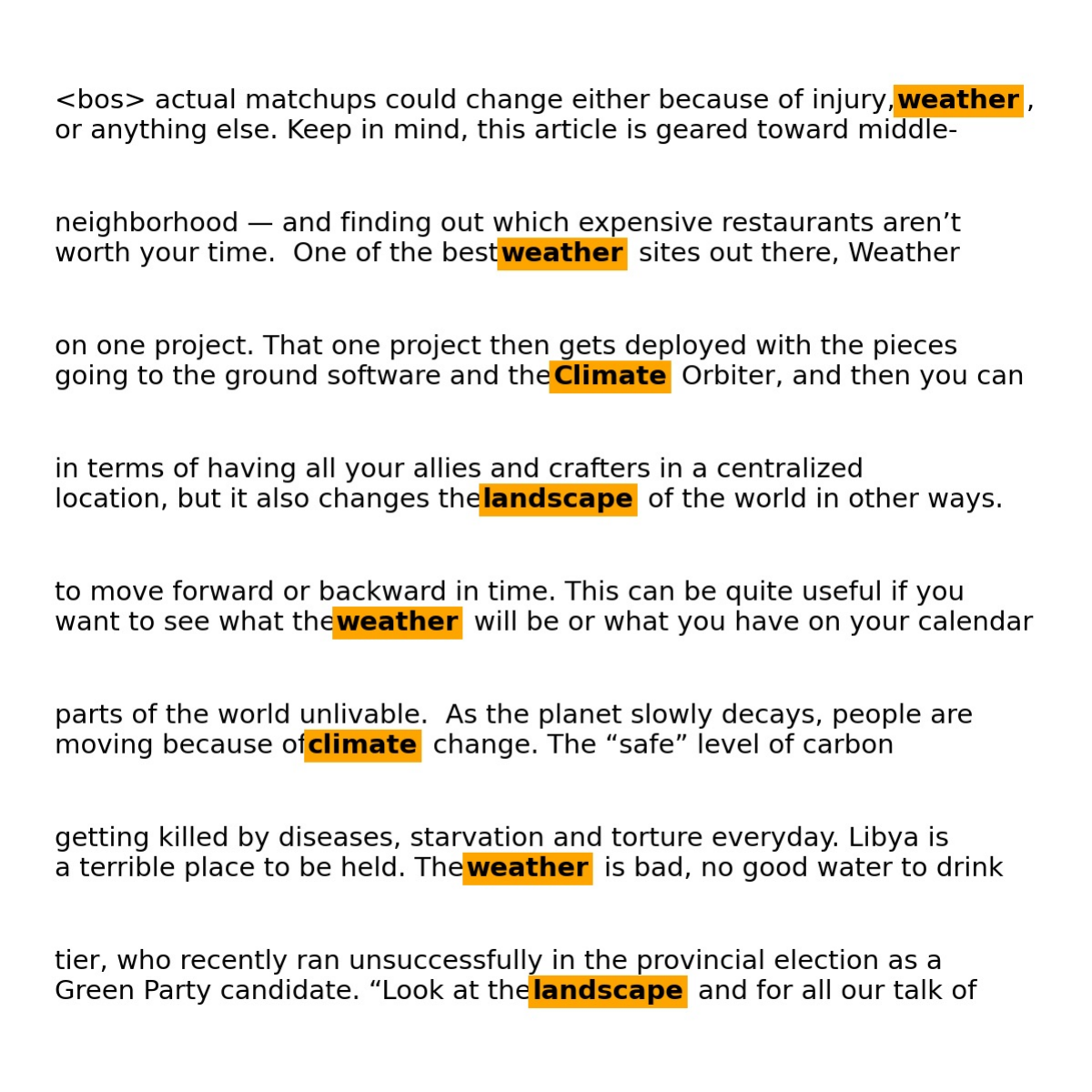}
        \caption{OrtSAE feature activating on broader environmental and weather contexts.}
        \label{fig:climate_2}
    \end{subfigure}
    \caption{\textbf{Decomposition of Climate-Related BatchTopK SAE Feature into OrtSAE Features.} (a) A BatchTopK SAE ($\text{L}0$=70) feature that activates on climate-related terms, which can be represented as a linear combination of two OrtSAE ($\text{L}0$=70): (b) An OrtSAE feature that activates specifically on climate-related terms. (c) An OrtSAE feature that activates on broader environmental and weather contexts.}
    \label{fig:climate_decomp}
\end{figure}

\begin{figure}[h!]
    \centering
    % First row - BatchTopK SAE feature
    % \begin{subfigure}[h]{0.25\textwidth}
    %     \centering
    %     \includegraphics[width=\textwidth]{./images/jaw_0.pdf}
    %     \caption{BatchTopK SAE feature activating on ``jaw'' token.}
    %     \label{fig:food_0}
    % \end{subfigure}
    
    % % Second row - Three OrtSAE features
    % \vspace{0.05cm}
    \begin{subfigure}[t]{0.245\textwidth}
        \centering
        \includegraphics[width=\textwidth]{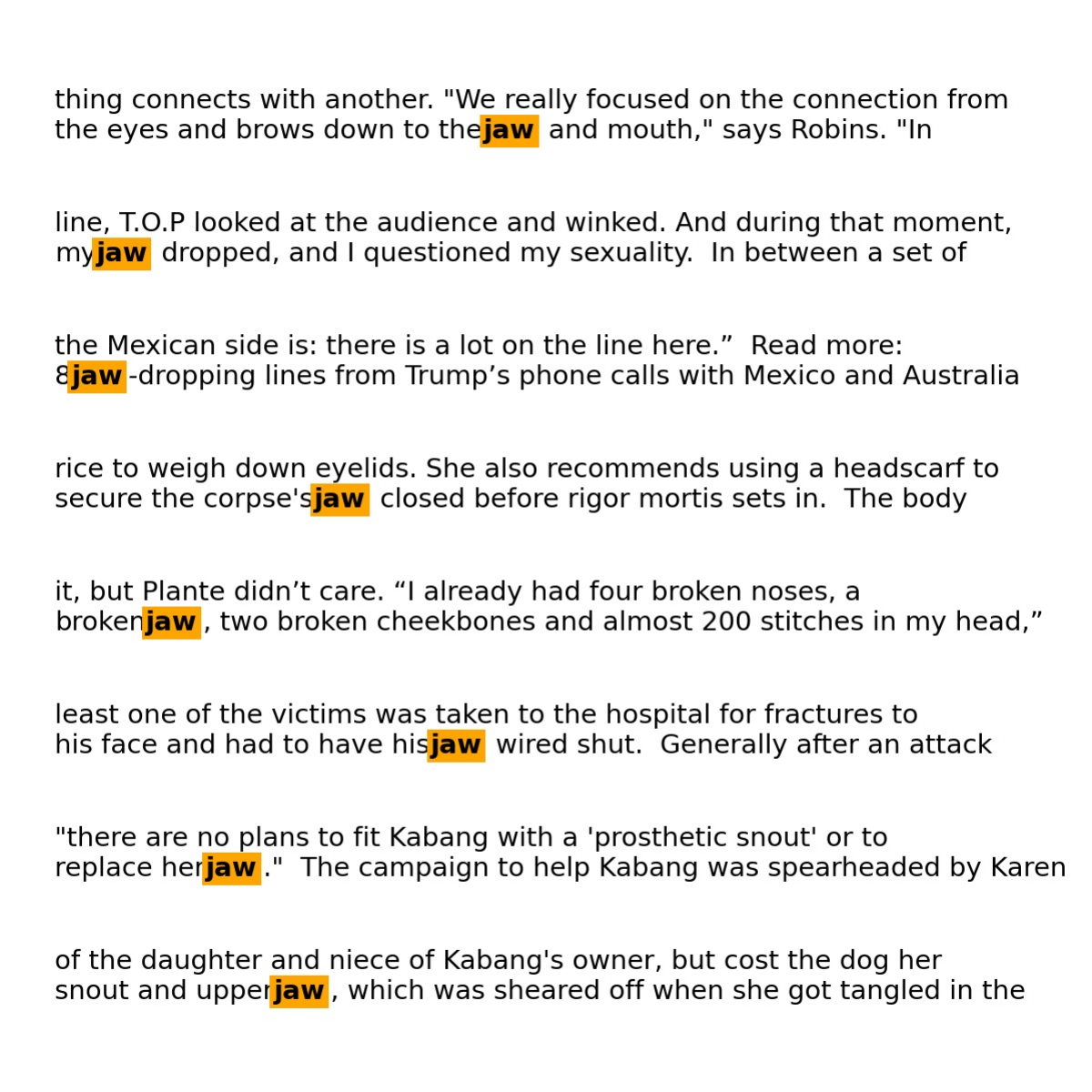}
        \caption{BatchTopK SAE feature activating on ``jaw'' token.}
        \label{fig:food_0}
    \end{subfigure}
    \begin{subfigure}[t]{0.245\textwidth}
        \centering
        \includegraphics[width=\textwidth]{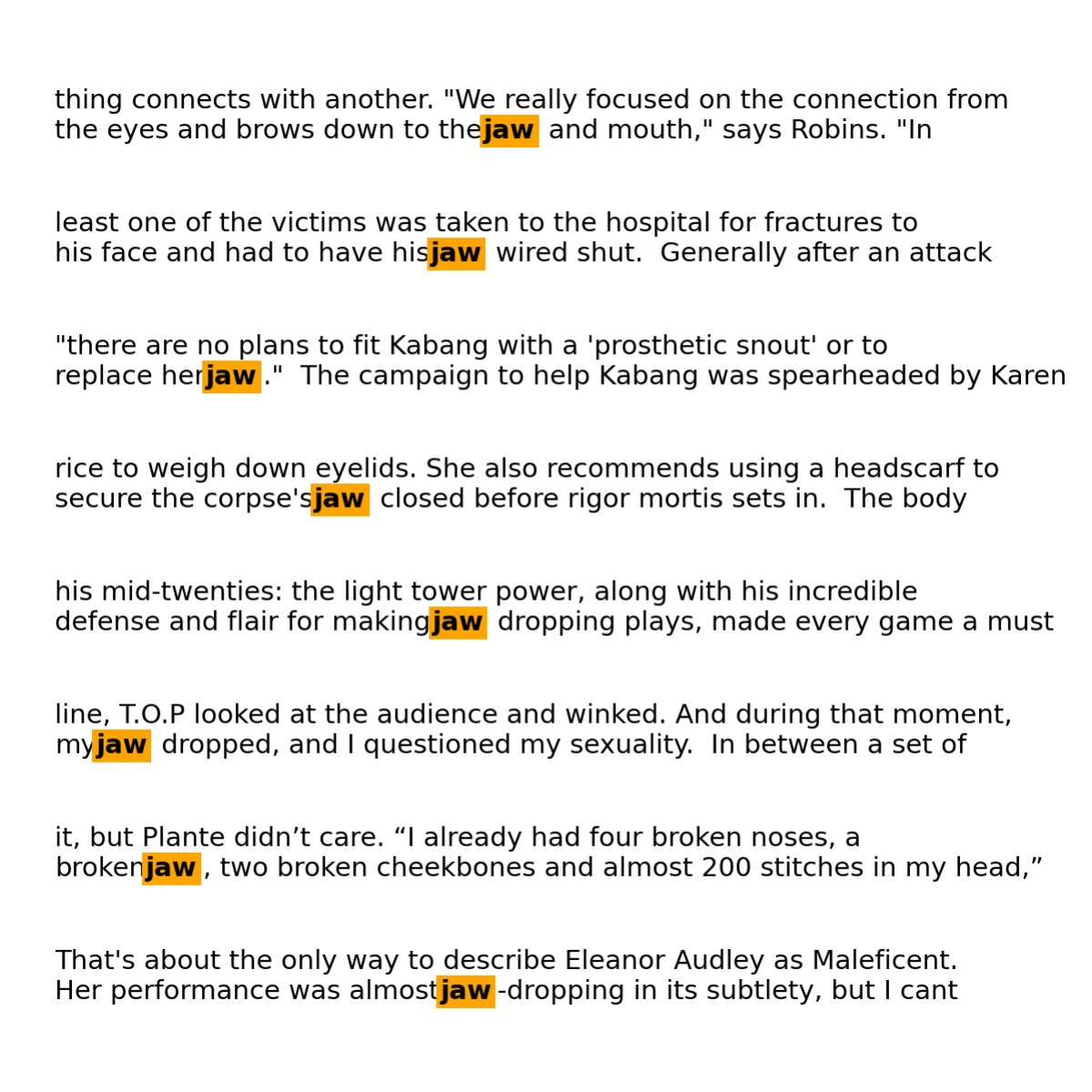}
        \caption{OrtSAE feature \\ activating on ``jaw'' token.}
        \label{fig:food_1}
    \end{subfigure}
    \begin{subfigure}[t]{0.245\textwidth}
        \centering
        \includegraphics[width=\textwidth]{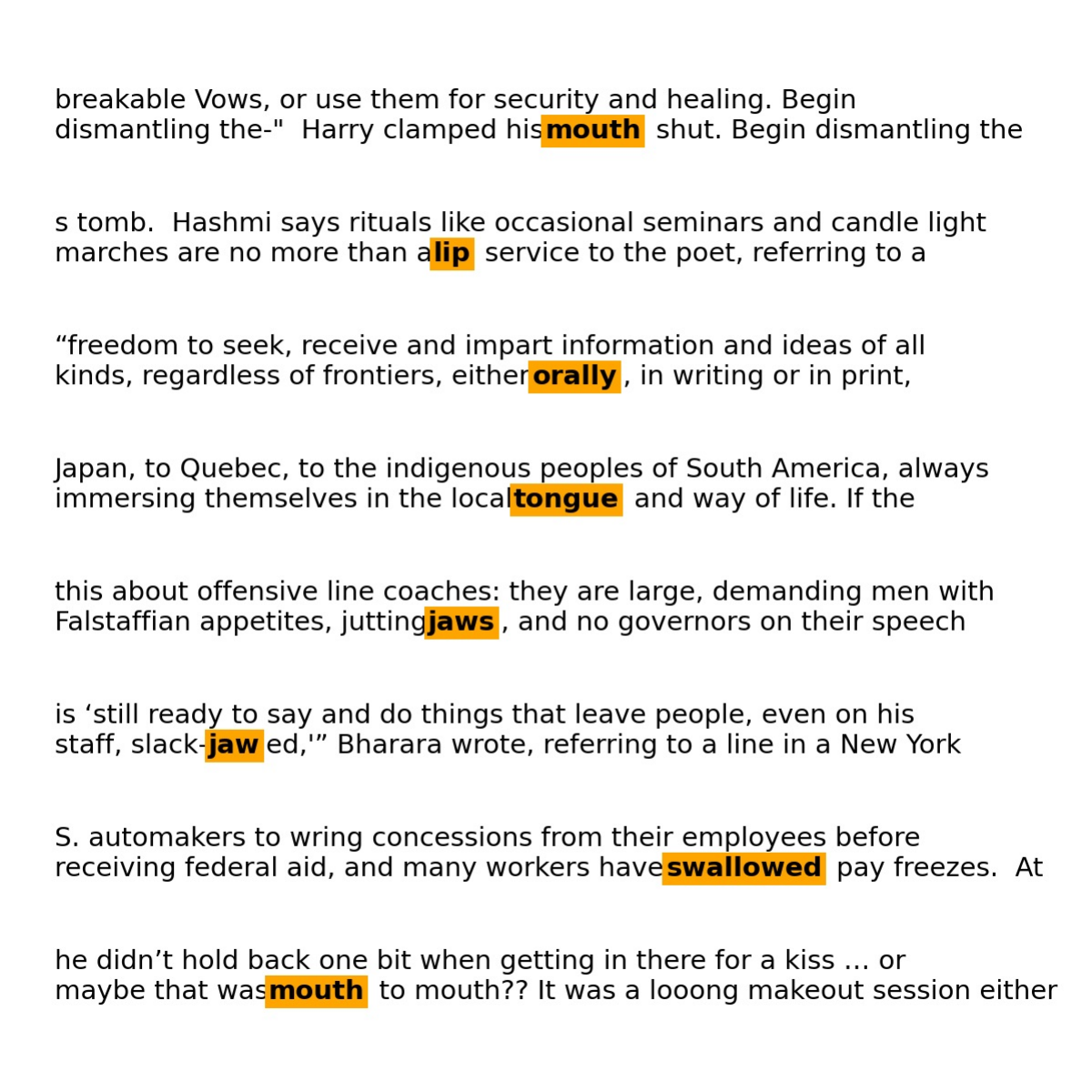}
        \caption{OrtSAE feature \\ activating on mouth \\ and oral concepts.}
        \label{fig:food_2}
    \end{subfigure}
    \begin{subfigure}[t]{0.245\textwidth}
        \centering
        \includegraphics[width=\textwidth]{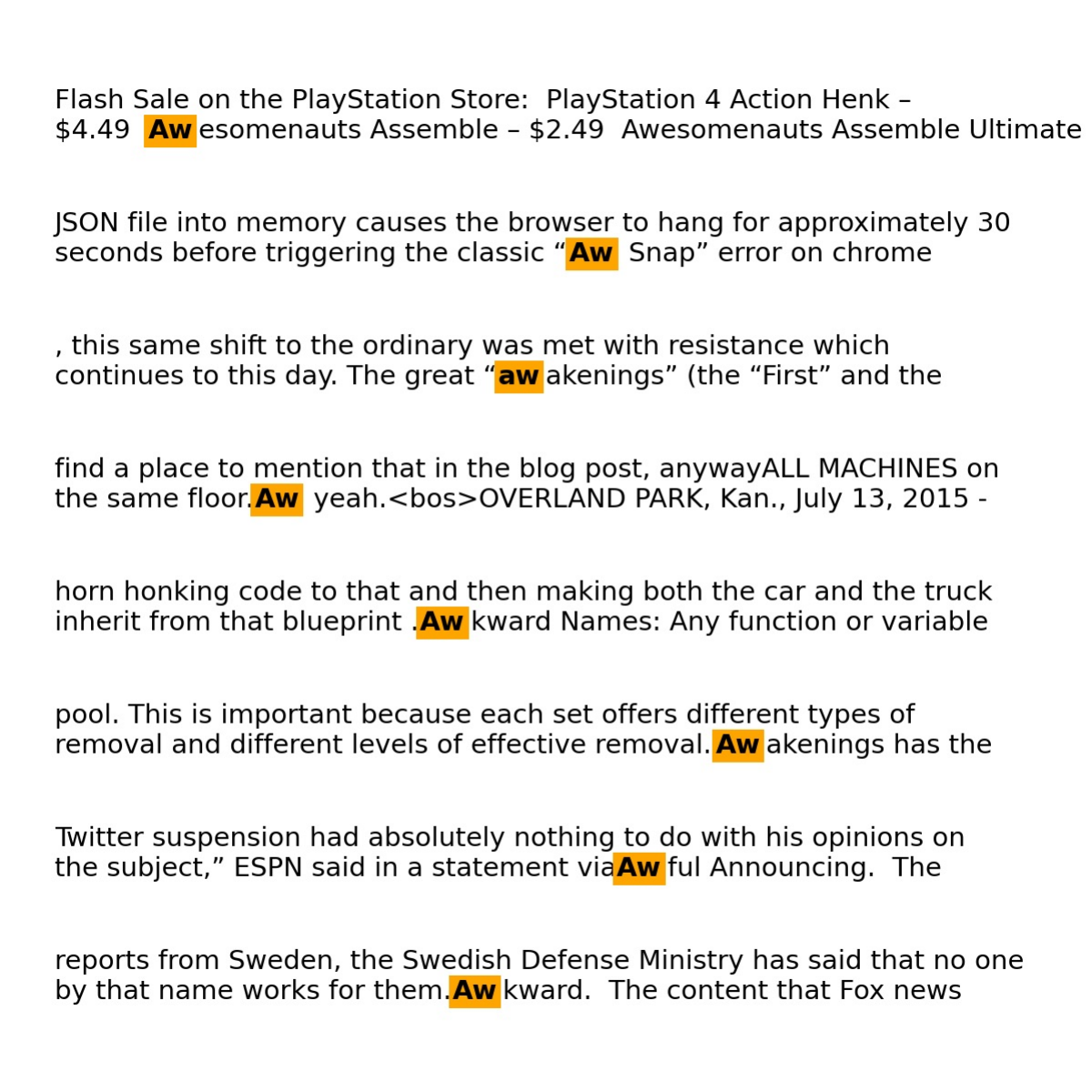}
        \caption{OrtSAE feature \\ activating on ``aw'' token.}
        \label{fig:food_3}
    \end{subfigure}
    \caption{\textbf{Decomposition of Jaw-Related BatchTopK SAE Feature into OrtSAE Features.} (a) A BatchTopK SAE ($\text{L}0$=70) feature that activates on the token ``jaw'', which can be represented as a linear combination of three OrtSAE ($\text{L}0$=70) features: (b) An OrtSAE feature that activates specifically on the token ``jaw''. (c) An OrtSAE feature that activates on mouth and oral concepts. (d) An OrtSAE feature that activates on the token ``aw''.}
    \label{fig:food_decomp}
\end{figure}

\begin{figure}[h]
    \centering
    % First row - BatchTopK SAE feature

    % Second row - Three OrtSAE features
    % \vspace{0.05cm}
    \begin{subfigure}[t]{0.245\textwidth}
        \centering
        \includegraphics[width=\textwidth]{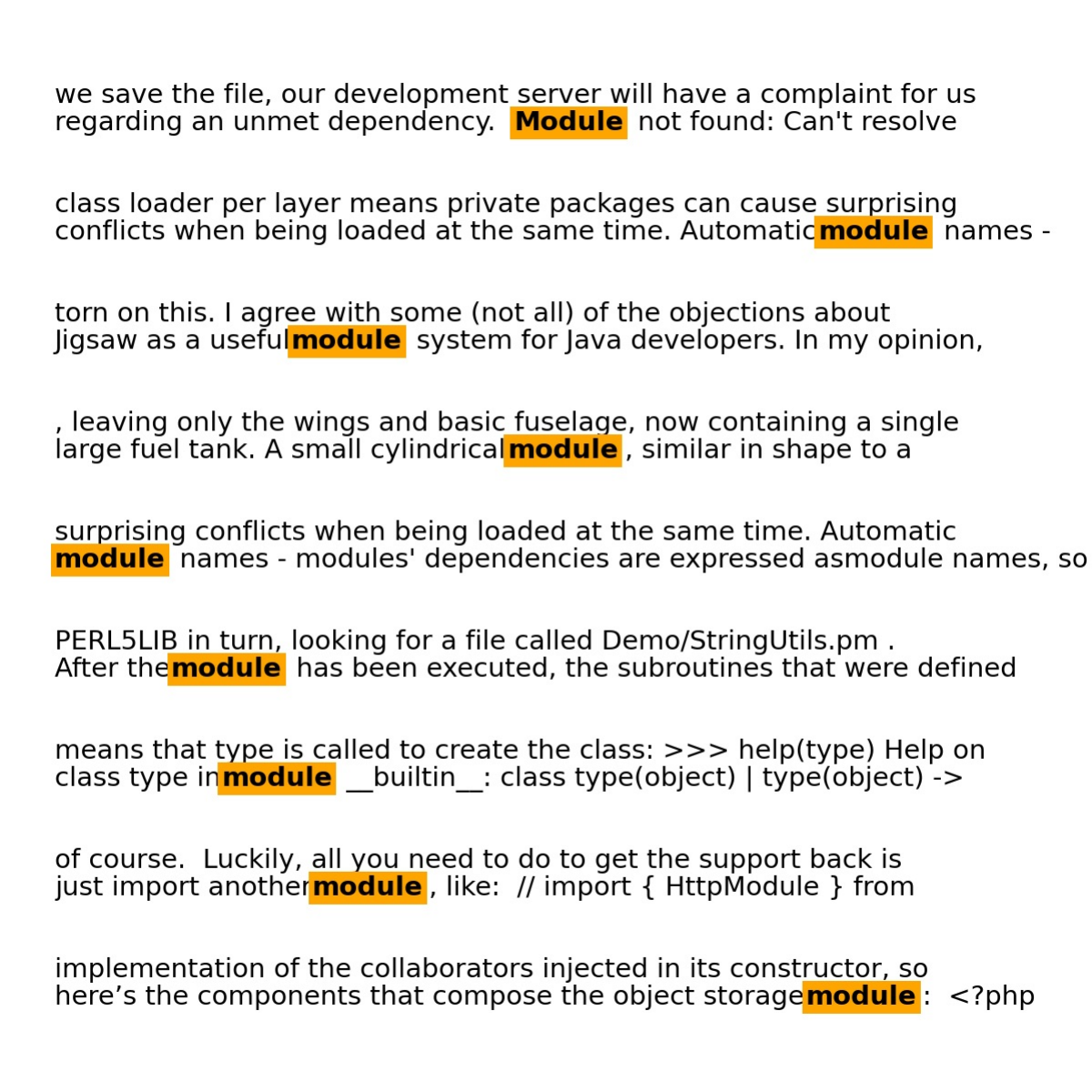}
        \caption{BatchTopK SAE feature activating on ``module'' \\ token.}
        \label{fig:tech_0}
    \end{subfigure}
    \hfill
    \begin{subfigure}[t]{0.245\textwidth}
        \centering
        \includegraphics[width=\textwidth]{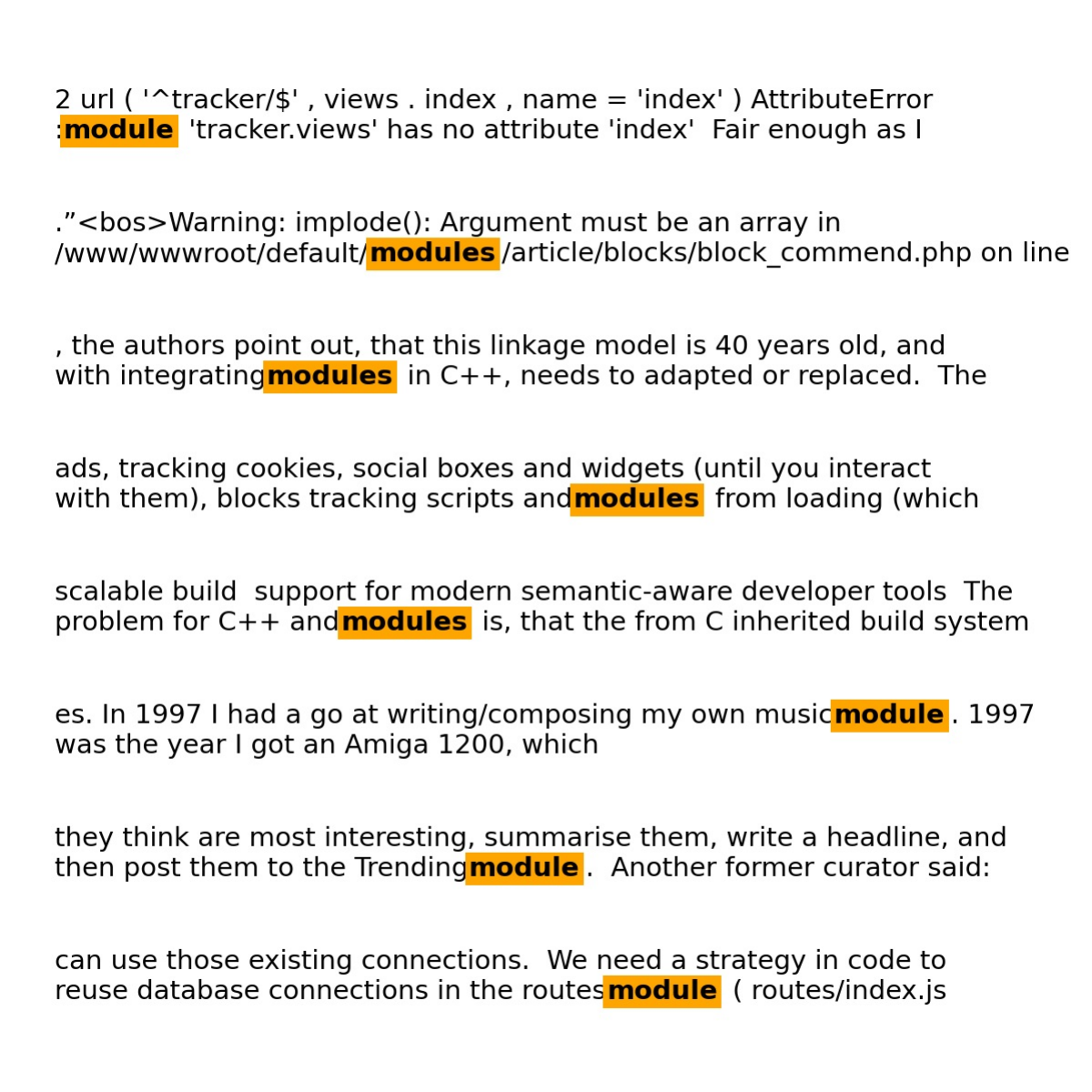}
        \caption{OrtSAE feature \\ activating on ``module'' \\ token.}
        \label{fig:tech_1}
    \end{subfigure}
    \hfill
    \begin{subfigure}[t]{0.245\textwidth}
        \centering
        \includegraphics[width=\textwidth]{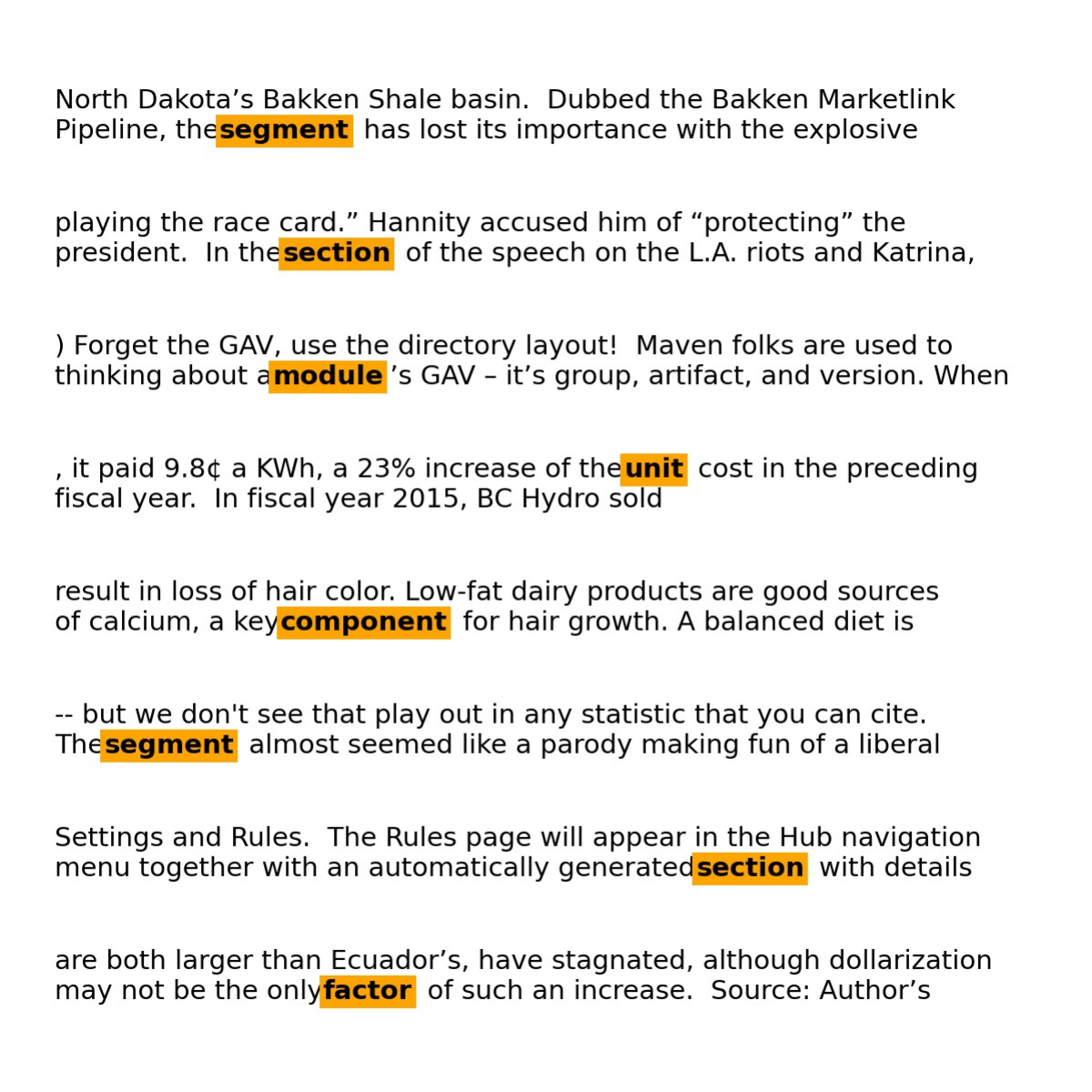}
        \caption{OrtSAE feature \\ activating on concepts \\ of parts, segments.}
        \label{fig:tech_2}
    \end{subfigure}
    \hfill
    \begin{subfigure}[t]{0.245\textwidth}
        \centering
        \includegraphics[width=\textwidth]{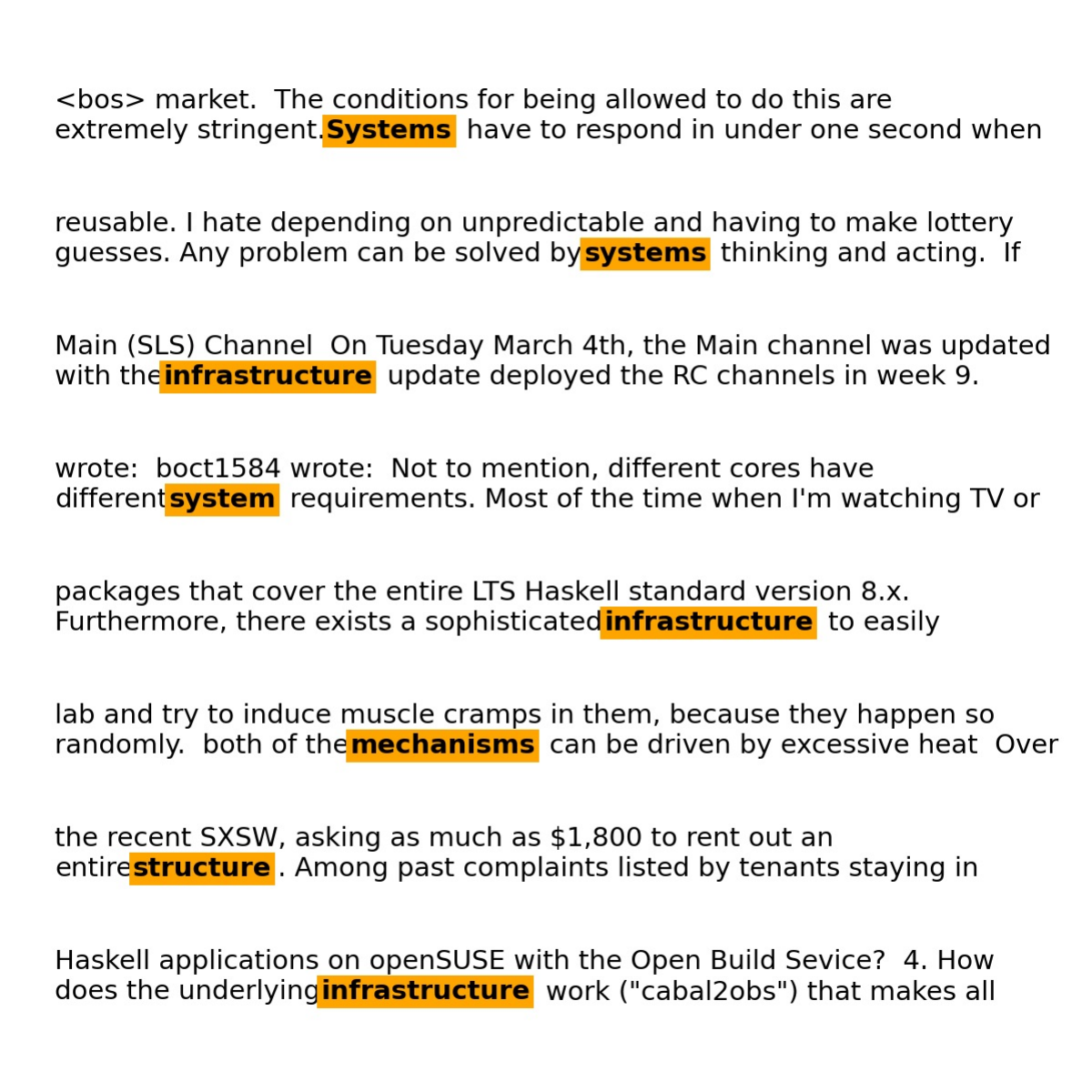}
        \caption{OrtSAE feature \\ activating on concepts \\ of complex system.}
        \label{fig:tech_3}
    \end{subfigure}
    \caption{\textbf{Decomposition of Module-Related BatchTopK SAE Feature into OrtSAE Features.} (a) A BatchTopK SAE ($\text{L}0$=70) feature that activates on the token ``module'', which can be represented as a linear combination of three OrtSAE ($\text{L}0$=70) features: (b) An OrtSAE feature that activates specifically on the token ``module''. (c) An OrtSAE feature that activates on concepts of parts and segments. (d) An OrtSAE feature that activates on concepts of complex systems.}
    \label{fig:tech_decomp}
\end{figure}

\end{document}